\documentclass{article} % For LaTeX2e
\usepackage{collas2024_conference,times}
\usepackage{comment}
\usepackage{booktabs}
\usepackage{subfigure}
\usepackage{subcaption}
\usepackage{diagbox}
\usepackage{adjustbox}
\usepackage{wrapfig}
\usepackage{multirow}
\usepackage{siunitx}
\usepackage{booktabs}
\usepackage{placeins}
\usepackage{threeparttable}

\usepackage{amsmath,amsfonts,bm}

\def\eqref#1{equation~\ref{#1}}
\def\1{\bm{1}}

\DeclareMathAlphabet{\mathsfit}{\encodingdefault}{\sfdefault}{m}{sl}
\SetMathAlphabet{\mathsfit}{bold}{\encodingdefault}{\sfdefault}{bx}{n}

\usepackage{hyperref}
\hypersetup{
    colorlinks=true,
    linkcolor=red,
    filecolor=magenta,
    urlcolor=blue,
    citecolor=purple,
    pdftitle={Overleaf Example},
    pdfpagemode=FullScreen,
    }

\usepackage[capitalize]{cleveref}
\crefname{section}{section}{sections}
\Crefname{section}{Section}{Sections}
\crefname{table}{Table}{Tables}
\Crefname{table}{Table}{Tables}
\crefname{figure}{Figure}{Figures}
\Crefname{figure}{Figure}{Figures}


\title{The Empirical Impact of Forgetting and Transfer\\in Continual Visual Odometry}

\author{Paolo Cudrano\thanks{Equal contribution.}\\ 
Politecnico di Milano\\
Milan, Italy \\
{\small\texttt{paolo.cudrano@polimi.it}}
\And
Xiaoyu Luo$^*$\\
Politecnico di Milano\\
Milan, Italy \\
{\small\texttt{xiaoyu1.luo@mail.polimi.it}}
\And
Matteo Matteucci \\
Politecnico di Milano\\
Milan, Italy \\
{\small\texttt{matteo.matteucci@polimi.it}}
}

\preprintcopy % Uncomment for the preprint version, but NOT for submission.

\begin{document}

\maketitle

\begin{abstract}
As robotics continues to advance, the need for adaptive and continuously-learning embodied agents increases, particularly in the realm of assistance robotics.
Quick adaptability and long-term information retention are essential to operate in dynamic environments typical of humans' everyday lives. A lifelong learning paradigm is thus required, but it is scarcely addressed by current robotics literature.
This study empirically investigates the impact of catastrophic forgetting and the effectiveness of knowledge transfer in neural networks trained continuously in an embodied setting. We focus on the task of visual odometry, which holds primary importance for embodied agents in enabling their self-localization. We experiment on the simple continual scenario of discrete transitions between indoor locations, akin to a robot navigating different apartments. 
In this regime, we observe initial satisfactory performance with high transferability between environments, followed by a specialization phase where the model prioritizes current environment-specific knowledge at the expense of generalization. 
Conventional regularization strategies and increased model capacity prove ineffective in mitigating this phenomenon. Rehearsal is instead mildly beneficial but with the addition of a substantial memory cost. 
Incorporating action information, as commonly done in embodied settings, facilitates quicker convergence but exacerbates specialization, making the model overly reliant on its motion expectations and less adept at correctly interpreting visual cues.
These findings emphasize the open challenges of balancing adaptation and memory retention in lifelong robotics and contribute valuable insights into the application of a lifelong paradigm on embodied agents.
\end{abstract}

\section{Introduction}
\label{sec:introduction}

The field of robotics has witnessed significant growth and has the potential to take a crucial role in various aspects of human life. An example of their potential is the domain of assistance robotics, where robots are designed to aid individuals in their daily activities, especially within home environments. For these robots to become integral components of our daily lives, it is essential that they not only engage with the specific environment in which they are originally deployed, but also that they adapt to it over time. We expect that the working environment is modified over time due to the user's daily activities. Moreover, the environment can change abruptly as users move from one place to another during their routine. It is crucial that a robotic system performs seamlessly across these changes.

Embodied agents, such as service robots, must thus possess the ability to continuously refine their capabilities in dynamic environments, a concept often referred to as lifelong learning. This notion also aligns with recent discussions,
such as those by \citet{betti2021can}, advocating for a move towards a lifelong learning paradigm, where machines acquire skills experiencing a human-like data stream and without the need for a pre-acquired database.
Within the realm of embodied robotics, topics such as memory, personalization, and enhanced interaction gain significance when considering adaptability after deployment. Unlike fixed, pre-trained models, adapting to the environment post-deployment allows for better customization and performance in real-world scenarios.

Despite the growing importance of these aspects, there is a noticeable gap in the existing literature, with a limited focus on robotics in lifelong scenarios \citep{LESORT202052}.  
On the contrary, ensuring that agents experience limited forgetting and demonstrate high knowledge transfer between different environments over time is key.

In this study, we empirically investigate these fundamental aspects in an initial scenario, focusing on the task of visual odometry. Given an agent that is free to move in an environment, visual odometry focuses on estimating its motion mainly using images acquired with a camera.
Building upon the groundwork laid by \citet{marullo2022continual}, who initially explored continual optical flow estimation from static video streams, our focus then shifts to embodied data streams and considers the implications of passive interaction with the environment, through motion.
Visual odometry holds notable importance for embodiment, as it addresses one of the agent's primary needs: self-localization. Differently from tasks traditionally studied in a continual setting, such as classification, detection, and segmentation of static images, visual odometry is influenced by external noise typical of robotics environments and is subject to a level of interaction with the environment, albeit passively through self-motion.

To characterize continual learning in an embodied scenario, we examine the degrees of possible environmental changes that an embodied agent can undergo and relate them to classical continual learning experience settings.
We then focus on the setting of discrete transitions between indoor locations over time. This setting reflects scenarios where a robot might navigate between different apartments, and it serves as a foundational step before addressing more complex real-world situations such as moving to offices, grocery stores, and other everyday environments.

We observe that through an initial stage of high forward transfer, the model achieves overall satisfactory performances when learning through different environments, acquiring environment-agnostic knowledge of the task at hand. After this initial convergence, however, a specialization phase takes place, leading to an over-enhanced performance in the current setting, at the expense of a significant reduction of generalization on past and future scenes.
This trade-off reflects the perpetual struggle between adaptation and retention of general features, a challenge that proves difficult to mitigate, even using conventional continual regularization strategies. Rehearsal provides slight mitigation, but at the cost of increased memory use and training time.
Interestingly, increasing model size, and thereby capacity, does not influence performance or modify this behavior. 

Following standard approaches from the embodied AI literature, we additionally provide the model with information about its intended motion action \citep{zhao2021surprising,memmel2023modality}. This information can help the model distinguish forward motions from rotations with ease, and since the agent is equipped with standardized discrete motion commands, it also provides exact information about the expected resulting motion.
Providing this additional information clearly improves performance. However, we find that while it highly enhances the learning convergence of the model, it also exacerbates its specialization to new environments. This action information makes the model more reliant on its knowledge of its standard motion primitives, and less adept at interpreting visual cues---a drawback particularly evident if actions fail to yield the desired outcomes, such as when encountering obstructions.

In summary, our study sheds light on the complexities of forgetting and transfer in neural networks deployed for visual odometry in embodied settings, underscoring the challenges associated with lifelong learning in robotics.
By focusing on a simple task and strategies, we aim to contribute valuable insights to the understanding of lifelong learning in robotics, specifically addressing the challenges posed by environmental changes and the delicate trade-off between adaptation and memory retention.

\section{Embodied AI}
\label{sec:embodied_ai}

Embodied AI studies systems that learn through their interaction with a physical environment, which can either be real or simulated. 
Learning becomes an interactive process between the agent and the environment, and takes place in an egocentric manner by interpreting information coming from the agent's sensors, possibly reacting to it through the agent's actuators.
This represents a paradigm shift from typical machine learning, which is instead based on the availability of large datasets to be consumed passively by the learning system all at once \citep{duan2022survey}.

Embodied systems face numerous challenges. As they operate in the real world, their environment is significantly complex, it is characterized by noise, and populated by other agents, each acting with potentially complex dynamics and, thus, hard to predict. Real-world objects are all characterized by a dynamic nature and are likely to change over time.
A long history of robotics literature has been testimony to these complexities for decades \citep{thrun2005probabilistic}. Throughout the years, these challenges have been repeatedly addressed and solved by focusing only on very restricted scenarios, solving specific problems in specific ways. The learning paradigm has only recently permeated the field, with works associating traditional techniques with deep learning frameworks. Nevertheless, the generated solutions are still limited.
Contrarily, the field of embodied AI, born from the deep learning community, focuses its attention on how to circumvent this specialization and produce data-driven solutions that are robust and more general.

As embodied AI advances, it expands the scope of the challenges it tackles. 
To interact with its environment, an agent generally needs to perceive its surroundings, know where it is located, and perform actions affecting itself and, possibly, the environment.
The development and testing of embodied agents in real-world conditions right from their infancy presents, however, several difficulties: on one hand, an untested system deployed in the real world, even in controlled conditions, might produce damage to itself and its environment, resulting in safety and economical risks; on the other hand, the complexities of dealing with physical hardware add to the problem technological issues, that are unrelated with the study of intelligent behaviors but can highly impact its progress and results. For this reason, in recent years the community has resorted to high-fidelity simulators, which have risen to very high standards to enable the study of high-level intelligent features.

Habitat-sim \citep{savva2019habitat} is one of the current state-of-the-art (SOTA) photo-realistic simulators, and allows to simulate the behavior of an agent equipped with RGB and depth sensors and moving through indoor locations. Subsequent iterations have also introduced object interactions, robotic arm manipulation, and the simulation of humanoid agents, thereby enabling the study of more intricate problems, such as object rearrangement and social interaction \citep{szot2021habitat,puig2023habitat}. An analogous photo-realistic simulator is iGibson \citep{xia2020igibson}, which supports the generation of high-quality virtual sensor signals, including RGB, depth, and LiDAR, and offers integration with purely robotics features, such as URDF support, controllers, and motion planners, facilitating complete interaction with environment objectives \citep{9636667}. AI2-THOR \citep{kolve2017ai2thor} offers also similar features, but is instead a game-based simulator, backed by Unity 3D, thus offering scalability in objects and scenes, at the price of reduced realism in its scenes.

In the embodied AI literature, the most studied tasks are visual exploration, visual navigation, and embodied QA \citep{duan2022survey}.
Most existing works in the field tackle these problems, however, with a standard learning mindset: given a large dataset representing the task, they learn how to best model it through a neural model.
This setup, however, assumes that the agent has access to a large amount of data before its deployment. This is possible during the initial development, but an additional learning mechanism must be in place for the agent to assimilate new information when it is already deployed.
For this reason, a different learning paradigm could be preferable \citep{betti2021can}.

An embodied agent is required to quickly adapt to its ever-changing environment, and be able to efficiently switch between different environments. This calls for a learning paradigm that improves the agent's abilities continually over time, in a lifelong manner. An embodied agent is efficient only if it shows high plasticity to new information and low forgetting of past data. Despite this consideration, not many works in the literature have explored this avenue \citep{marullo2022continual}.

\vspace{-.5em}
\subsection{Visual Odometry}
\begin{figure}[t]
\centering
\begin{minipage}{0.5\linewidth}
  \centering
  % \includesvg[width=\linewidth]{img/embodied_vo_setup.svg}
  \includegraphics[width=\linewidth]{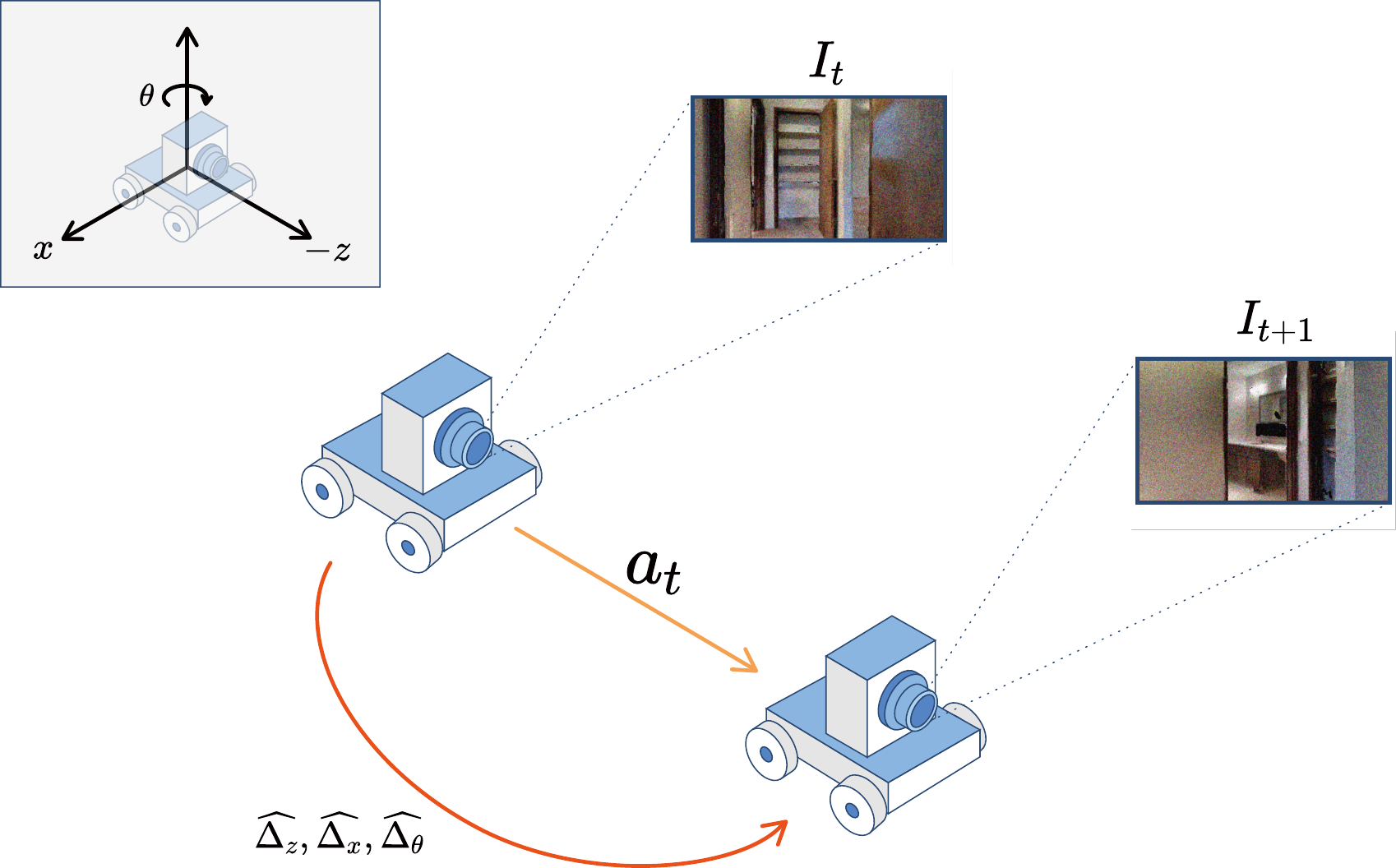}
\end{minipage}
\hspace{0.07\linewidth}
\begin{minipage}{0.35\linewidth}
  \centering
  \includegraphics[width=\linewidth]{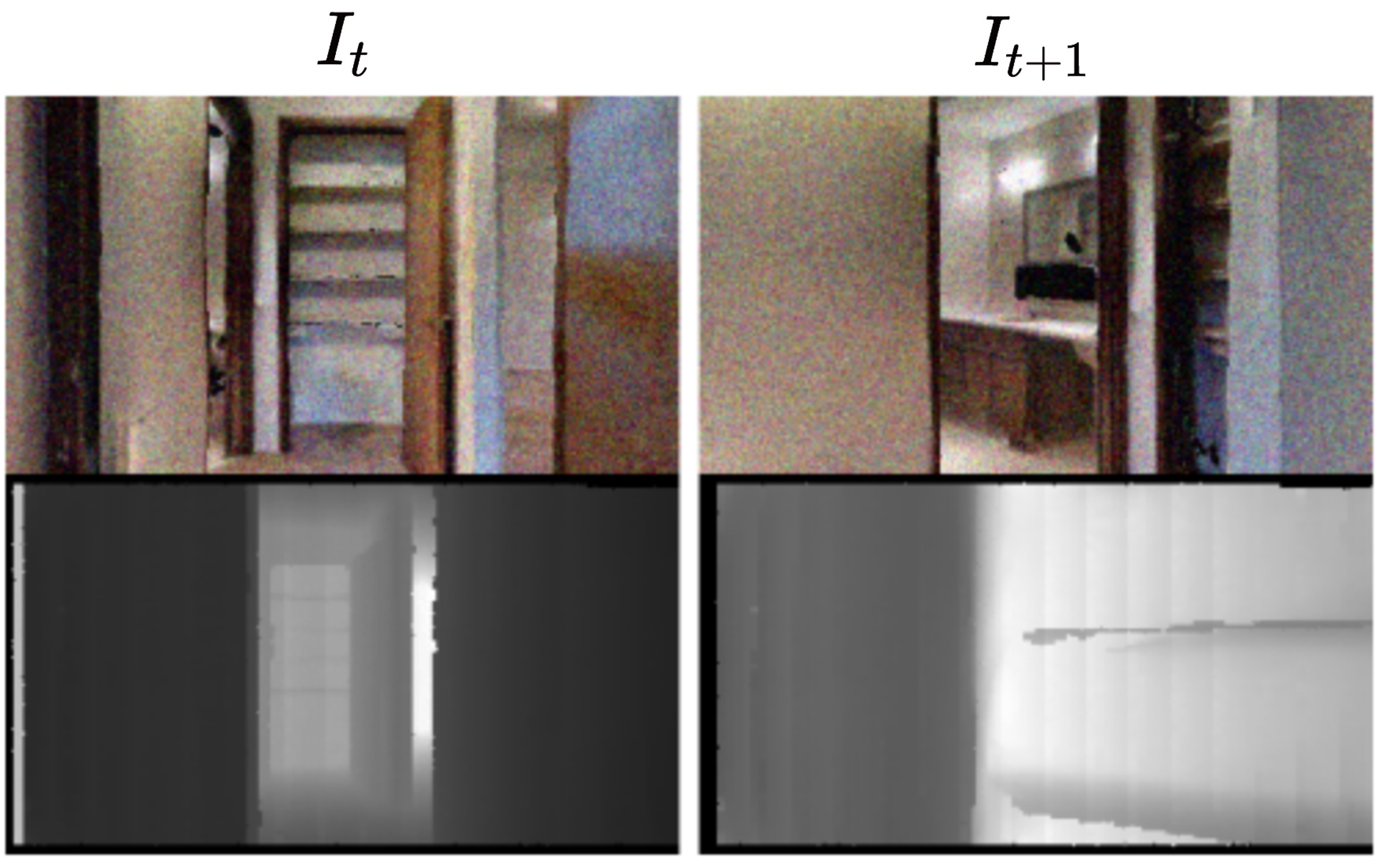}
\end{minipage}
  \caption{\textbf{Visual odometry (VO)}. A mobile agent equipped with a camera is in motion in its environment while observing the scene. As the agent is subject to noisy actuation, the actual motion of the vehicle is not known at high precision after a motion action $a_t$ is performed. Visual odometry focuses on estimating the actual displacement $(\widehat{\Delta_z}, \widehat{\Delta_x}, \widehat{\Delta_{\theta}})$ registered by the agent in a short time step $t+1$, using the information acquired via cameras at each time step ($I_t, I_t+1$).
  }
  \label{fig:vo}
  \vspace{-1em}
\end{figure}
To perform tasks in an embodied setting, an agent
must satisfy several basic needs. Among these, agents must be capable of knowing or estimating their own positions within the environment, in order to navigate it, recognize what is around them, and interact with other objects. This ability to accurately track the agent's movements is even more essential in indoor environments, where GNSS and other positioning systems are not available or practical.

As typically assumed in embodied AI, we can consider the high-level motion of the robot as being the composition of a few primitive actions, such as \textit{moving forward}, \textit{turning left}, or \textit{turning right}. By doing so, we are assuming that the physical control of the robot is managed at a lower level, while we can focus on the intelligent behavior only.
We point out however that in the physical world, it is never possible to control everything, and that noise is always present. In robotics, particularly, external factors such as ground conditions, wheel friction, and environmental forces, as well as internal factors such as battery state, hardware delays, and software scheduling, contribute to perturbing the final outcome of a motion command. 
To counteract this well-known phenomenon in robotics, 
it is important to estimate how much the robot actually moved after every motion command.

Visual odometry (VO) consists exactly in estimating the robot's displacement after a brief unit of time, particularly using vision sensors.
Typical visual inputs are images from one or more RBG cameras. Additional depth estimation can be performed using stereovision or adopting specific RGB-D sensors. Information from other proprioceptors is sometimes included, such as from inertial systems (IMU). Additional information about the most recent motion command executed is also often used to aid the estimation. \Cref{fig:vo} depicts the typical visual odometry setup.

While traditional VO algorithms relied on handcrafted features, recent research has focused on deep neural networks to learn VO directly from raw images. These data-driven methods offer the ability to learn complex motion patterns and adapt to various environments. Notable works include \citet{zhou2017unsupervised} and \citet{clark2017vinet}, who proposed end-to-end trainable VO systems using deep CNNs and recurrent neural networks, respectively.
\citet{yang2020d3vo} introduced D3VO, integrating deep depth, pose, and uncertainty estimation into a direct VO model. \citet{zhu2022deepavo} introduce a four-branch network that leverages CNNs to focus on different quadrants of the optical flow input for learning rotation and translation. Lastly, \citet{memmel2023modality} introduce a Visual Transformer \citep{dosovitskiy2020image} based VO, achieving Modality-Invariant performance in the presence of multiple sensors.

\section{Continual Learning in Embodied Environments}
\vspace{-1em}
The dynamic nature of embodied environments is naturally suitable for learning continually. Indeed, physical worlds present lots of interacting components and, even with large-scale datasets, it is hard to capture most of its variability when building a generalist agent.
To complicate things, part of this variability can also be due to the agent's behavior.
It can be argued, indeed, that the agent should learn mostly through direct interaction with its environment \citep{smith2005development}. In this regard, we can identify and define two levels of interaction:
\begin{description}
\item[Passive interaction.] The agent interacts with the environment only through its own motion. The agent's motion has a direct effect on how the agent perceives its environment, but does not alter the environment in any way.
\item[Active interaction.] The agent is not only free to move, but it also directly changes the state of its environment (e.g., moving or grasping objects).
\end{description}

It must be noted that when interacting with the physical world, objects break or wear out, and the interaction can occasionally fail. At the same time, the agents themselves may encounter malfunction or hazardous states. When engaging in active interaction, for instance, agents may inadvertently cause damage or encounter unforeseen consequences, some of which can be destructive. 
For this reason, it is crucial to provide a learning environment that minimizes risks. 
Additionally, the fixed time dynamics of the physical world may limit the speed at which a trial-and-error learning process can be performed.
To address these limitations, as highlighted in \Cref{sec:embodied_ai}, researchers have turned to simulations, where interactions cannot cause catastrophic consequences and the learning process can be sped up in time \citep{duan2022survey}.
Indoor environments present unique challenges to be captured by simulators. Indeed, indoor spaces are typically confined, lack global navigation satellite system (GNSS) signals, and are populated with humans, all of which add to the complexity of the tasks the agents need to perform.

In the context of traditional continual learning, researchers categorize the learning problem in three scenarios \citep{van2019three}: domain-incremental, task-incremental, and class-incremental learning.
In domain-incremental learning, the task remains constant but the data distribution changes over time; in task-incremental learning, both the tasks and the data distribution change over time, but the set of classes remains constant; and in class-incremental learning, new classes are also introduced over time, thereby increasing the complexity of the tasks. We realize that this categorization, although essential in classical continual learning, does not consider the complexity introduced by an embodied environment.
In particular, we identify an additional problem dimensionality, which has to do with the granularity of the considered continual learning scenario. We define this granularity, as follows, focusing on the problem of visual odometry, but we believe the principles remain general:
\begin{description}
\item[Scene-level.] The agent learns a task within a single apartment (or scene) at a time, avoiding forgetting when moving forward. The agent can collect and store as many trajectories in the apartment before learning.
\item[Trajectory-level] 
As the agent visits different apartments over time, each trajectory represents a distinct learning experience, and past trajectories do not remain available. The agent is allowed to shuffle observations within a single trajectory, breaking their strong temporal correlation.
\item[Observation-level] The agent learns from each new data sample over time, and cannot revisit past observations. This is equivalent to a pure online setup.
\end{description}

\section{Continual VO: Experimental Setup}
\vspace{-1em}
We work in indoor environments, which are typical for most service robots, such as home helpers, and can pose specific challenges to agents. Indeed, in these environments, agents cannot exploit GNSS-based self-localization, and their motion must be curated to avoid collisions due to the confined space.
We exploit the quality and diversity of scenes provided by Habitat simulator \citep{savva2019habitat}, which contains a large set of 72 photorealistic apartments derived from real-world data. In Habitat, the agent is a robot equipped with an RGB-D sensor. Additional information about its localization $\left(x, y, \theta \right)$ can be accessed, and it is usually treated as ground truth. As the agent moves around the apartment following trajectories from point A to point B, the agent collects camera images, its poses, and records the actions performed at each timestep. The available actions are discrete: forward, rotate right, and rotate left. Motion noise is also modeled as a combination of Gaussian noise, typically imputable to imperfect control or feedback sensors, and sharper disturbances due to collisions with obstacles such as furniture and walls.

Our objective is to study how learning the task of visual odometry in this embodied frame can lead to potentially different transfer and forgetting properties. 
To do so, we focus on the most simple scenario, to avoid potential confounders in our analysis.
We consider the \textit{passive interaction} originating from the agent's motion, which we assume to be provided by an optimal policy. We notice that, although the agent does not actively modify the environment, its perception changes significantly, and the i.i.d. hypothesis vanifies.
We consider a \textit{scene-level granularity}, where the agent is immersed in sequentially new apartments over time, similarly to what a potential robot helper would experience when following a human throughout its daily activities.

We base our work on the VO model by \citet{zhao2021surprising}, known in the embodied AI literature. 
We consider their dataset, which is collected through Habitat by repeatedly sampling a trajectory in one of the 72 Habitat apartments. 
Each sample is composed of a couple of 4-channels RGB-D images, $x_t, x_{t+1}$, with each image $x = (x_R,x_G,x_B,x_D)$. Additionally, the last executed action $a_t$ can also be recorded. The actions provided by Habitat are fixed: move forward $a_{fwd}$, rotate left $a_{left}$ and rotate right $a_{right}$. Details on their average motion can be found in \cref{sec:action_prediction_variability}.
Together with this information, the ground-truth displacement is collected, as $\mathbf{\Delta} = (\Delta_z, \Delta_x, \Delta_{\theta})$. In Habitat conventions, the $z$ axis points towards the back of the robot, the $x$ axis points to its left, and the yaw is measured counterclockwise (\cref{fig:vo}).
Notice that, although the policy used is optimal, the agent might still collide with parts of the environment during its trajectory, leading to samples particularly out-of-distribution.
As in traditional learning works, \citeauthor{zhao2021surprising} shuffle all samples across trajectories and scenes, rendering the distribution effectively i.i.d..
As we cannot utilize this technique in a continual setup, we exploit their data collection pipeline to obtain an analogous dataset in which each apartment, or experience, is represented independently, and samples are not shuffled. We then split this dataset into train, validation, and test sets, with quota $ 82 \%, 6 \%, 12 \% $. \Cref{sec:dataset} further describes the composition of our dataset, highlighting the balance between each action and reporting the number of collisions registered.

We build on the model presented in \citet{zhao2021surprising}. The authors use an adapted Resnet\nobreakdash-18 \citep{he2016deep} with 3.94M parameters and train it on the RGB-D data for regressing the displacement of the agent $\mathbf{\Delta} = (\Delta_z, \Delta_x, \Delta_{\theta})$ occurred in the last time step $\delta_t$. 
In their work, they perform ablations on the impact of the D channel, as well as on adding information about the control action just performed. 
While the RGB images alone provide only visual features, the depth channel D can aid in relating image to world features by disentangling the representation from the camera calibration. Knowing the particular action, moreover, the model can confidently predict values close to the expected motion, using the image only to slightly compensate for noise.
The RGB and D information are classically encoded as 3- and 1- channel images respectively.
The model receives as input the RGB-D frames for $t$ and $t+1$, which are concatenated along the channel dimension, resulting in inputs of size $341 \times 192 \times 8$. 
\citeauthor{zhao2021surprising} deal with the action information by building 3 separate backbones---one for each possible action---, trained separately at first and combined later with a particular augmentation strategy.
We find this approach inefficient, as for each new action it requires a new model, which would then become extremely specialized on the particular motion distribution due to such action.
For this reason, when studying the impact of the action information, we avoid duplication and maintain a single model. We exploit the action information as an additional input, concatenating it along the channel dimension so that it can be accessed homogeneously by every convolution filter.

Following the original work \citep{zhao2021surprising}, we train our model with MSE regression loss:
\begin{equation}
\label{eq:loss}
\mathcal{L} = \frac{1}{N} \sum_{i=1}^{N} \left((\Delta_{x,i} - \widehat{\Delta_{x,i}})^2 + (\Delta_{z,i} - \widehat{\Delta_{z,i}})^2 + (\Delta_{\theta,i} - \widehat{\Delta_{\theta,i}})^2\right),
\end{equation}
where $N$ is the number of samples considered, $ ( \Delta_z,\Delta_x,\Delta_\theta )$ is the ground truth agent pose, and $ ( \widehat{\Delta_z},\widehat{\Delta_x},\widehat{\Delta_\theta} )$ is the predicted pose of the agent. For convenience, we define also:
\begin{align}
\label{eq:loss_components}
\mathcal{L}_z = \frac{1}{N} \sum_{i=1}^{N} \left|\Delta_{z,i} - \widehat{\Delta_{z,i}} \right|, \qquad
\mathcal{L}_x = \frac{1}{N} \sum_{i=1}^{N} \left| \Delta_{x,i} - \widehat{\Delta_{x,i}} \right|, \qquad
\mathcal{L}_\theta = \frac{1}{N} \sum_{i=1}^{N} \left| \Delta_{\theta,i} - \widehat{\Delta_{\theta,i}} \right|.
\end{align}

Starting from a random initialization, we train the network continually over all 72 apartments provided in the Habitat train set. For each apartment, also referred to as experience, we train our model for a maximum of 40 epochs, with early stopping on the validation set. We use Adam optimizer, batch size 32, and learning rate \num{2e-4}.
We additionally train a joint baseline on the entire shuffled dataset of 72 apartments, maintaining the number of epochs to 40, and using batch size 128 and learning rate \num{2e-4}.
The test set is extracted from a separate set of 14 unseen apartments, provided in the Habitat evaluation set.

\section{Forgetting and Transfer}
\label{sec:forgetting_and_transfer}
We analyze whether learning VO continually by interacting with a single environment at a time determines a decrease in performance when moving to new environments, i.e., forgetting. Complementarily, we study whether during training the capabilities acquired in one scene improve the overall performance on past scenes (backward transfer) or aid in learning future scenes (forward transfer).

\begin{figure}[t]
\hspace*{\fill}%
\begin{minipage}[t]{.45\textwidth}
  \centering
  \vspace{0pt}
  \includegraphics[width=\linewidth]{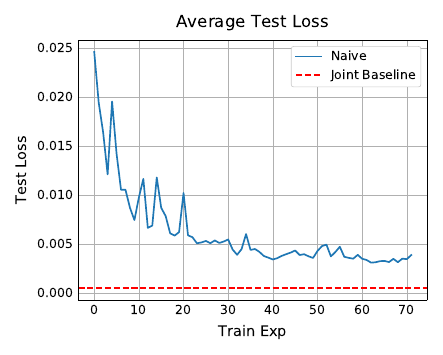}
  \vspace{-2em}
  \captionof{figure}{Average test loss across all experiences when performing naive finetuning, i.e., training on the sequence of apartments continually. We compare against a joint training on all apartments, noticing a large gap in the converged performance of the two methods (\num{3.88e-3} against \num{0.50e-3}).}
  \label{fig:avg_loss}
\end{minipage}%
\hfill
\begin{minipage}[t]{.46\textwidth}
  \centering
  \vspace{0pt}
  \includegraphics[width=\linewidth]{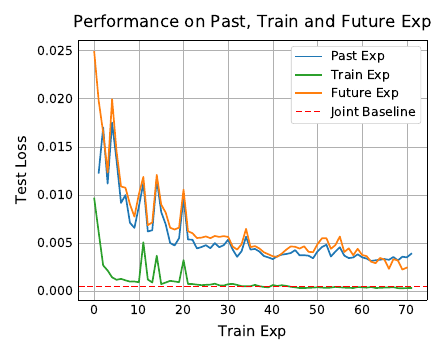}
  \vspace{-2em}
  \captionof{figure}{Progression during lifelong training of the loss reached on the current apartment against the loss scored on past or future apartments. The graph highlights how, after an initial phase, the network improves only on the apartment it is currently visiting, while past and future experiences remain flat.}
  \label{fig:loss_pres_past_fut}
\end{minipage}
\hspace*{\fill}%
\vspace{-.5em}
\end{figure}

To perform this evaluation, after training on each experience, we measure the test loss on every experience in the dataset (i.e., current, past, and future).
We can then plot in \Cref{fig:avg_loss} the overall average test loss behavior throughout the life of our model. We notice that the average test loss decreases throughout training, indicating that the model acquires further general knowledge of the task while progressing through the experiences. However, as time passes, the improvement decreases and converges to a flat region.

\begin{figure}[t]
\begin{center}
    \subfigure
    % {\includegraphics[width=0.33\textwidth]{img/res18_naive_dz.eps}}
    {\includegraphics[width=0.33\textwidth]{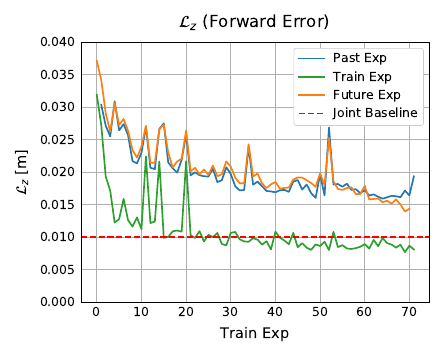}}
    \subfigure
    % {\includegraphics[width=0.33\textwidth]{img/res18_naive_dx.eps}}
    {\includegraphics[width=0.33\textwidth]{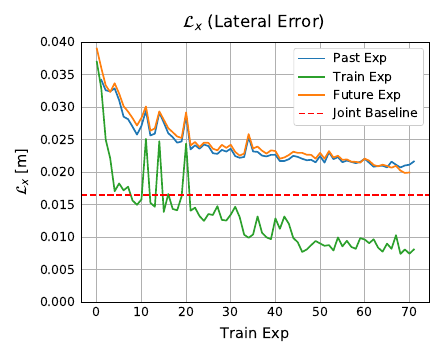}}
    \subfigure
    % {\includegraphics[width=0.32\textwidth]{img/res18_naive_dyaw.eps}} \\
    {\includegraphics[width=0.32\textwidth]{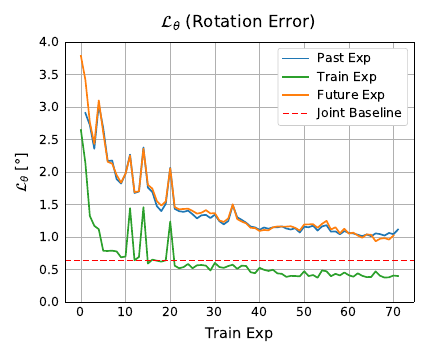}} \\
\end{center}
\vspace{-1.5em}
\caption{Comparison of each loss component over current, past, and future experiences.}
\label{fig:loss_pres_past_fut_separate}
\vspace{-.5em}
\end{figure}

If we analyze the performance separately on the current experience and compare it with all past and future ones, we notice a different trend (\Cref{fig:loss_pres_past_fut}). The first few experiences provide a general improvement on the overall task, but after about 30 experiences, we only see improvements in the performance on the current experience, which decreases significantly, while the loss on other experiences remains stable. This indicates that, after a certain time, the model focuses only on improving its current behavior, but does not acquire further transferable knowledge on the general task. This effect is even more evident when considering the three output components $(\Delta_z, \Delta_x, \Delta_{\theta})$ separately (\Cref{fig:loss_pres_past_fut_separate}).

\begin{figure}[t]
\begin{center}
    \subfigure
    % {\includegraphics[width=0.34\textwidth]{img/res18_naive_bwt.eps}}
    {\includegraphics[width=0.34\textwidth]{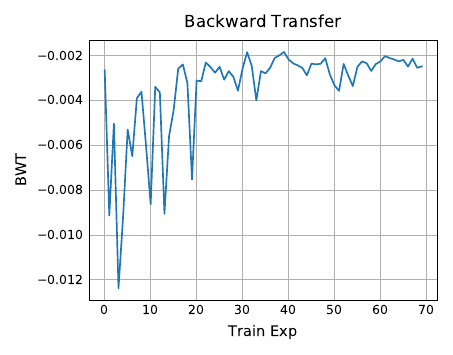}}
    \subfigure
    % {\includegraphics[width=0.31\textwidth]{img/res18_naive_forgetratio.eps}}
    {\includegraphics[width=0.31\textwidth]{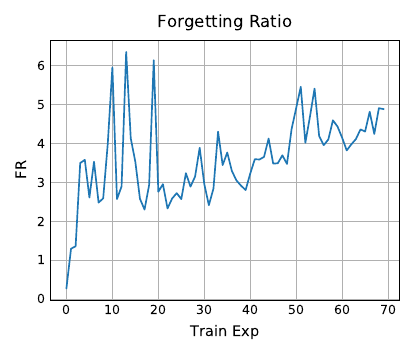}}
    \subfigure
    % {\includegraphics[width=0.33\textwidth]{img/res18_naive_fwt.eps}}
    {\includegraphics[width=0.33\textwidth]{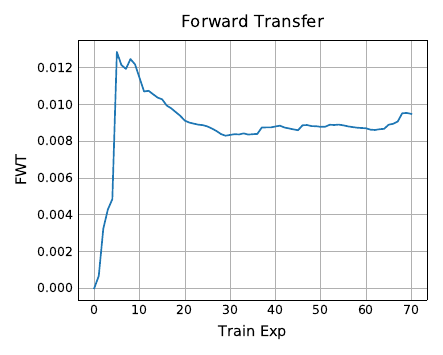}}
\end{center}
\vspace{-1.4em}
\caption{Comparison of continual learning metrics: backward transfer, forgetting ratio, and forward transfer.\vspace{-.5em}}
\label{fig:forgetting_and_transfer}
\end{figure}

We can further study this behavior using typical continual learning metrics backward transfer, forward transfer \citep{lopez2017gradient}, and forgetting ratio \citep{he2020continuous}.
Considering $\mathcal{L}_{k,j}$ as the loss measured on experience $j$ after training on experience $k$, we can then redefine these classical metrics in the context of regression as follows:
\begin{itemize}
    \item Backward transfer (BWT), measures how much current learning benefits also past experiences, 
    i.e., the improvement that learning experience $k$ brings on average to past experiences $j < k$,
    \begin{equation}
        \operatorname{BWT_k} = \frac{1}{k-1} \sum_{j=1}^{k-1} \left( \mathcal{L}_{j,j} - \mathcal{L}_{k,j} \right).
    \end{equation}
    \item Forgetting ratio (FR), quantifies forgetting on past experiences in terms of relative decrease from the ideal value, i.e., the average relative loss decrease registered on past experiences $j$ when learning experience $k$,
    \begin{equation}
        \operatorname{FR_k} = \frac{1}{k-1} \sum_{j=1}^{k-1} \frac{\max{\left(0, \mathcal{L}_{k,j} - \mathcal{L}_{j,j} \right)}}{\mathcal{L}_{j,j}}.
    \end{equation}
    \item Forward transfer (FWT) measures how much learning benefits future experiences through transfer, i.e., how much all past experiences $j$ improved learning for the current experience $k$:
    \begin{equation}
        \operatorname{FWT_k} = \frac{1}{k-1} \sum_{j=2}^{k} \left( \tilde{\mathcal{L}_{j}} - \mathcal{L}_{j,j} \right),
    \end{equation}
    where $\tilde{\mathcal{L}_{j}}$ is the loss obtained training a model from scratch on experience $j$.
\end{itemize}
We report these metrics in \Cref{fig:forgetting_and_transfer}. We can see how positive forward transfer is present throughout the experiment, as the model steadily exploits its knowledge to improve its performance on new experiences. However, in doing so, it also experiences increasing forgetting, meaning that to perform better on new experiences, it has to consistently lose its edge on past scenes, thus never achieving fully general capabilities across the entire suite of apartments. This is reflected in an increasing forgetting ratio and an always-negative backward transfer.

Given the poor results obtained on our naive baseline, we proceed to evaluate the effectiveness of mitigation techniques well-known in the continual learning literature, focusing first on regularization-based methods, and then on rehearsal methods.

\subsection{Regularization Strategies}

\begin{figure}[!t]
\begin{minipage}{0.49\linewidth}
    \centering
    \includegraphics[width=0.9\linewidth]{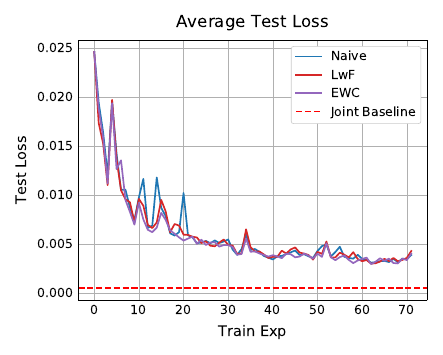}
    \vspace{-1em}
    \caption{Impact of continual regularization strategies.}
    \label{fig:regularization}
\end{minipage}
\hfill
\begin{minipage}{0.49\linewidth}
\centering
\captionsetup{type=table} % tell latex to change to table
\caption{Average Test Loss}
\label{tab:performance_summary}
\begin{threeparttable}
    \centering
    \begin{tabular}{l | SS }
        \toprule
        \textbf{Experience} & \textbf{Average} & \textbf{Final}\\
        \midrule
        Naive           & 6.23 & 3.88 \\
        EWC             & 6.05 & 4.29\\
        LwF             & 5.86 & 4.03 \\
        Replay (1024)   & 5.48 & 3.39 \\
        Replay (5120)   & 4.67 & 2.69 \\
        Replay (13888)  & 3.99 & 2.26 \\
        \midrule
        \multicolumn{2}{l}{Joint Baseline} & 0.50 \\
        \bottomrule
    \end{tabular}
    \begin{tablenotes}
      \small
      \item All loss values expressed in scale $\num{e-3}$.
    \end{tablenotes}
\end{threeparttable}
\end{minipage}
\end{figure}

Among the strategies proposed in the continual learning literature, regularization techniques are especially desirable for embodied tasks, as they do not require additional memory or compute consumptions.
In \cref{fig:regularization}, we explore the impact of the continual regularization strategies EWC \citep{kirkpatrick2017overcoming} and LwF \citep{li2017learning}. \cref{sec:apx_regularization} reports other continual learning metrics.
Overall, our results show that the trend of VO loss and continual metrics presents the same behavior obtained with traditional finetuning, suggesting that it is not straightforward to mitigate the issue with regularization techniques.

\subsection{Rehearsal Strategies}

As regularization techniques do not mitigate forgetting in continual VO, we resort to the family of rehearsal-based methods. These methods require the storage of a limited amount of past information in extra memory, which could be composed of actual copies of past samples, or be more elaborate, such as being a deep generator model trained to mimic past data.

To explore the benefits introduced by strategies based on rehearsal, we evaluate the change in performance when introducing a random replay of past experiences, storing some of the past samples in a limited buffer. We evaluate buffer sizes of 1024, 5120, and 13888 (respectively about 10\%, 50\%, and 100\% of the data in one apartment). We select the samples to store randomly, still guaranteeing an equal proportion of samples for each past apartment. 

\begin{figure}[!htp]
\hspace*{\fill}%
\begin{minipage}[t]{.45\textwidth}
  \centering
  \vspace{0pt}
  \includegraphics[width=0.9\linewidth]{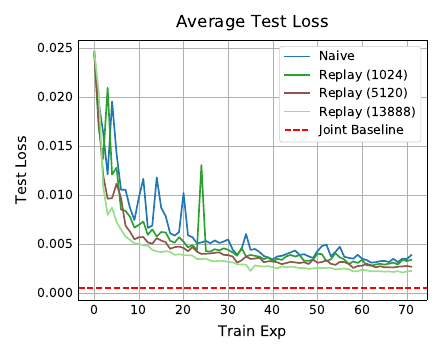}
  \vspace{-.5em}
  \captionof{figure}{Effect of rehearsal-based methods, highlighting the impact of different buffer sizes (about 10\%, 50\%, and 100\% of a single training apartment).}
  \label{fig:res18_replay}
\end{minipage}%
\hfill
\begin{minipage}[t]{.45\textwidth}
  \centering
  \vspace{0pt}
  \includegraphics[width=\linewidth]{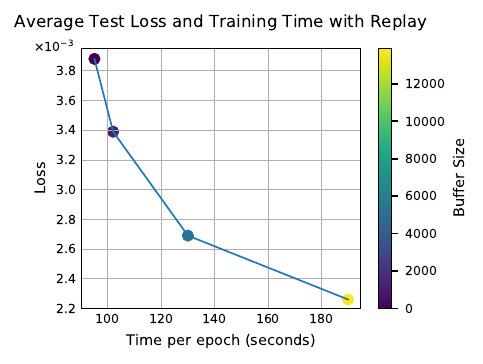}
  \captionof{figure}{Increasing replay buffer size leads to better performance at the cost of higher training times.}
  \label{fig:2d_buffer_time}
\end{minipage}
\hspace*{\fill}%
\end{figure}

As depicted in \cref{fig:res18_replay}, rehearsal methods do have some effect in mitigating the overall forgetting. Expectedly, their impact is proportional to the size of the buffer used. However, even when the buffer contains as many samples as the current experience, the improvement is very limited. Moreover, when evaluating on past, current, and future apartments (\cref{sec:apx_reharsal}), no clear winner between all replay methods could be found. \Cref{tab:performance_summary} summarizes our results, reporting the average validation loss of each method over all 72 experiences. We report the loss obtained at the end of training (Final) and the average loss obtained after training on each new experience (Average). \cref{sec:apx_reharsal} includes further figures for completeness.

Increasing the buffer size, the amount of sample seen by the model in one epoch increases, and so does the training time. We highlight this phenomenon in \cref{fig:2d_buffer_time}, pointing out that larger buffers achieve better performance at the cost of longer training times, and suggesting that a tradeoff analysis must be carried out during deployment.

\section{Impact of Action Information}

Because of the deeper challenges often encountered in embodied scenarios, works in embodied AI tend to provide as much information as possible to the agent to improve its performance.
In VO, it is common practice to assist the model also with information on the last control action performed, as this is typically known by the system and strongly correlates with the actual motion of the agent.
If the action space is discrete, like in our case, this datum conveys a lot of information to the model and allows it to learn the expected output for each different action individually. This considerably simplifies the task, as the RGB-D images need to be used only to reassess the impact of noise on each single motion command.
Notice that, formally, this corresponds to learning only
\begin{equation}
p(\mathbf{\Delta} | x_{rgb}, a ),
\end{equation}
which is significantly easier than predicting the entire motion only from the RGB-D data. In this case, indeed, the target distribution would present multiple modes:
\begin{equation}
p(\mathbf{\Delta} | x_{rgb} ) = \sum_{a \in A} p(\mathbf{\Delta} | x_{rgb}, a ) \cdot p(a ).
\end{equation}
Given its practical relevance, we study how knowledge of the action performed can affect the performance and forgetting behavior of the network. We provide this action information to the model by encoding it as an additional input channel, assuming constant value (${+1,0, \text{or} -1}$).

\begin{figure}[htp]
\centering
\begin{minipage}[t]{.47\textwidth}
  \centering
  \vspace{0pt}
  \includegraphics[width=.98\linewidth]{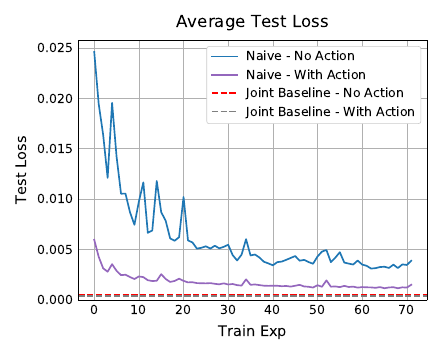}
  \captionof{figure}{Average test loss across all experiences when the model is fed action information.}
  \label{fig:avg_loss_action}
\end{minipage}%
\hfill
\begin{minipage}[t]{.47\textwidth}
  \centering
  \vspace{0pt}
  \includegraphics[width=\linewidth]{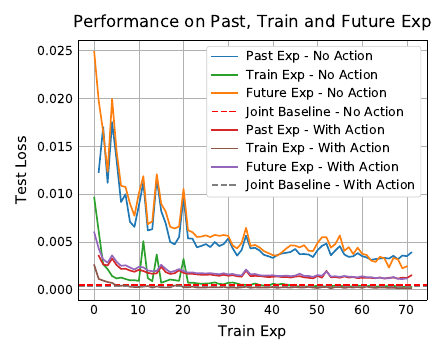}
  \captionof{figure}{Comparison of the loss over current, past, and future experiences when the model is fed action information.}
  \label{fig:loss_pres_past_fut_action}
\end{minipage}
\end{figure}

As done in \cref{sec:forgetting_and_transfer}, we report in \cref{fig:avg_loss_action,fig:loss_pres_past_fut_action} the overall average test loss, together with the loss computed only over the past, current and future experiences.
Having access to the action significantly improves general performance not only when jointly training, as also reported by \citet{zhao2021surprising}, but also in a continual setting. We further notice that not only the model achieves lower errors, but its convergence is sped up significantly. We conjecture that this happens because the model can focus on a single motion distribution at a time, instead of having also to determine which particular motion has been performed directly from the images.
We notice that a substantial performance gap is present when compared to the action-less model. Despite attempts, we have not been able to bridge it by increasing compute, data, or model capacity. 
We report further analyses in \cref{sec:action_prediction_variability}.

\section{Impact of Model Scale}\label{sec:impact_of_scale}

We investigate the impact of the model size on performance and forgetting behavior. Up to this point, our study has been conducted with a ResNet18 adopted from the embodied AI literature, which is typically parsimonious on computational requirements to function on lightweight devices. Nevertheless, we question whether the small scale of the model can in general impact the forgetting and transfer of a VO model, as indicated also by \citet{ramasesh2021effect}.

\begin{figure}[t]
\hspace*{\fill}%
\begin{minipage}[t]{.47\textwidth}
  \centering
  \vspace{0pt}
  \includegraphics[width=.98\linewidth]{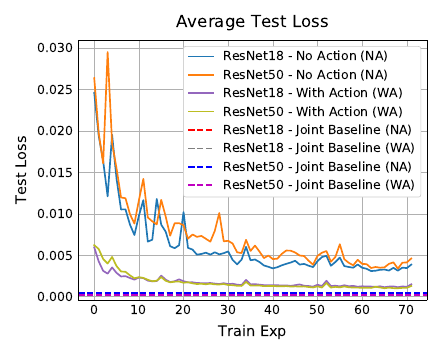}
  \captionof{figure}{Impact of scaling the model on the average test loss across all experiences.}
  \label{fig:avg_loss_scale}
\end{minipage}%
\hfill
\begin{minipage}[t]{.47\textwidth}
  \centering
  \vspace{0pt}
  \includegraphics[width=\linewidth]{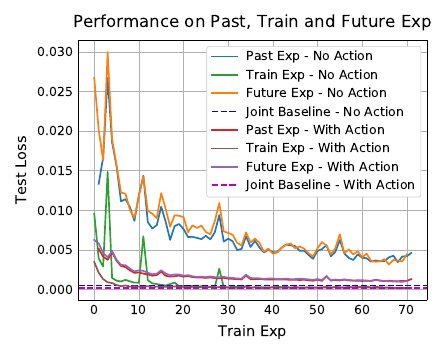}
  \captionof{figure}{Comparison of the loss over current, past, and future experiences for a larger Resnet50 model, showing the same trends.}
  \label{fig:loss_pres_past_fut_scale}
\end{minipage}
\hspace*{\fill}%
\end{figure}

We adopt a significantly larger model, based on Resnet50 \citep{he2016deep} and thus having more than $6 \times$ more parameters. We train this model in the same setup used in previous sections, obtaining a joint baseline and a continually trained model over all 72 experiences.
The results, shown in \cref{fig:avg_loss_scale,fig:loss_pres_past_fut_scale}, strongly suggest that increasing the model size has no relevant impact on the model's forgetfulness and transfer in the context of this study.
The experiments confirmed, also in this case, the absence of improvements when applying regularization techniques EWC and LwF. An analysis of the prediction variability, presented in \cref{sec:action_prediction_variability}, additionally confirms our previous findings.

\section{Discussion and Limitations}

In this study, we investigated the impact of continual learning (CL) in embodied environments, focusing on visual odometry (VO) with a scene-level granularity. Our findings shed further light on the challenges and limitations of applying CL to embodied AI tasks.

We observed that, even with the most relaxed granularity and with only passive interaction, the embodied task of continual VO exhibits signs of forgetting. This phenomenon persists despite attempts to mitigate it through simple regularization techniques or increasing the model capacity. Rehearsal strategies partially reduce the phenomena but require additional computational costs and large memory buffers at training time. 

These results highlight the inherent difficulties of adapting CL to embodied settings, where the agent must contend with additional challenges introduced by its immersion in a physical environment.
Our findings underscore the importance of considering the unique requirements of embodied AI when developing CL approaches. Moving forward, it is essential to explore alternative CL techniques that may better address the challenges posed by embodiment. Our results raise questions on whether the standard CL setup for non-embodied contexts is sufficient to tackle the lifelong learning challenges inherent to embodied environments.

While this study may provide valuable insights, we realize that our analysis is limited to a specific model architecture and scenario, and further research is needed to generalize our findings across different settings. Exploring the efficacy of other CL techniques in embodied contexts remains also an important avenue for future investigations.

In conclusion, our study contributes to the growing body of research at the intersection of CL and embodied AI, highlighting the need for tailored approaches to address the unique challenges posed by embodiment. By highlighting new dimensions of the continual learning problem when adopted in embodied scenarios, we aim to bridge the gap between robotics and continual learning, advancing the development of intelligent agents capable of thriving in real-world settings over time.

\subsubsection*{Acknowledgments}
This paper is supported by the FAIR (Future Artificial Intelligence Research) project, funded by the NextGenerationEU program within the PNRR-PE-AI scheme (M4C2, investment 1.3, line on Artificial Intelligence).

\bibliography{collas2024_conference}
\bibliographystyle{collas2024_conference}

\newpage % added

\appendix
\section{Additional Evaluations}

In the main paper, we report only the results most significant to our analysis. We include here further results for completeness, highlighting trends analogous to what can be found in the main paper.

\subsection{Regularization Strategies}
\label{sec:apx_regularization}

\begin{figure}[!htp]
\begin{center}
    \subfigure
    % {\includegraphics[width=0.33\textwidth]{img/res18_stg_res50_bwt.eps}}
    {\includegraphics[width=0.33\textwidth]{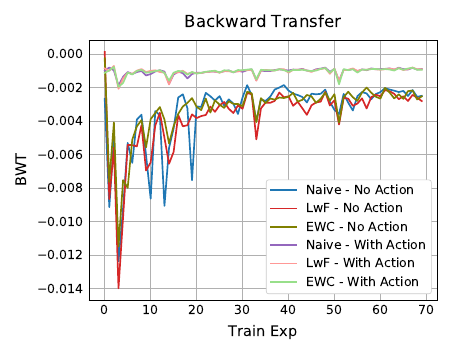}}
    \subfigure
    % {\includegraphics[width=0.3\textwidth]{img/res18_stg_res50_ratio.eps}}
    {\includegraphics[width=0.3\textwidth]{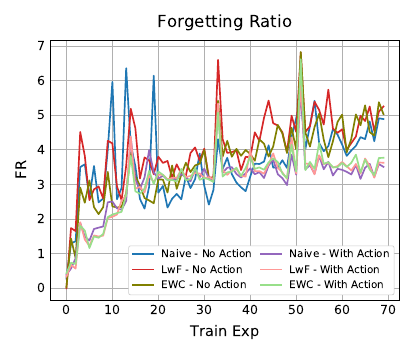}}
    \subfigure
    % {\includegraphics[width=0.32\textwidth]{img/res18_stg_fwt.eps}} \\
    {\includegraphics[width=0.32\textwidth]{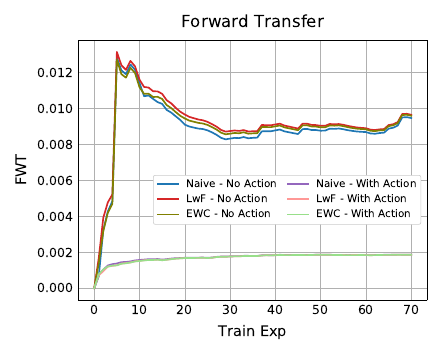}} \\
\end{center}
\vspace{-1em}
\caption{Continual metrics, computed on regularization strategies, show that EWC and LwF are as ineffective as the naive baseline in mitigating forgetting or improving transfer.}
\label{fig:cl_metrics_reg}
\end{figure}

\subsection{Rehearsal Strategies}
\label{sec:apx_reharsal}
\begin{figure}[!htp]
\begin{center}
    \subfigure
    % {\includegraphics[width=0.33\textwidth]{img/buffer1024_loss.eps}}
    {\includegraphics[width=0.33\textwidth]{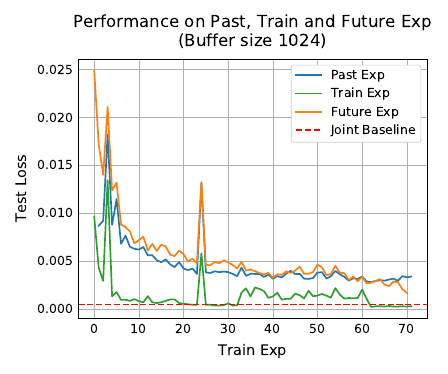}}
    \subfigure
    % {\includegraphics[width=0.33\textwidth]{img/buffer5120_loss.eps}}
    {\includegraphics[width=0.33\textwidth]{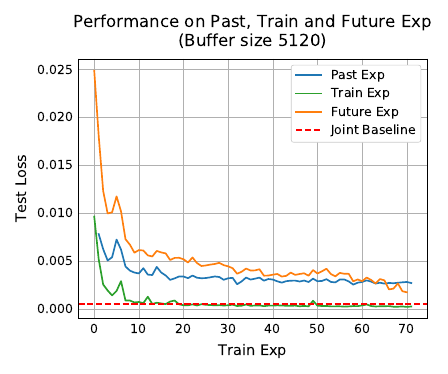}}
    \subfigure
    % {\includegraphics[width=0.33\textwidth]{img/buffer13888_loss.eps}}
    {\includegraphics[width=0.33\textwidth]{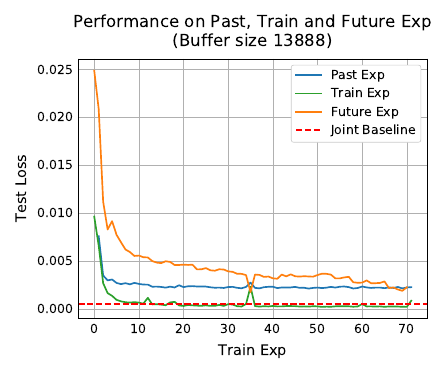}}
\end{center}
\vspace{-1em}
\caption{Impact of replay (using different buffer sizes) on the loss over current, past, and future experiences.}
\label{fig:replay_different_buffer}
\end{figure}
\begin{figure}[!htp]
\begin{center}
    \subfigure
    % {\includegraphics[width=0.33\textwidth]{img/replay_bwt.eps}}
    {\includegraphics[width=0.33\textwidth]{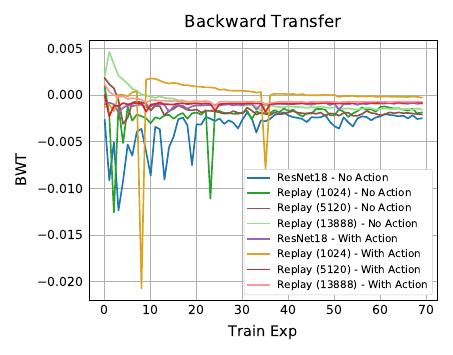}}
    \subfigure
    % {\includegraphics[width=0.32\textwidth]{img/replay_fr.eps}}
    {\includegraphics[width=0.32\textwidth]{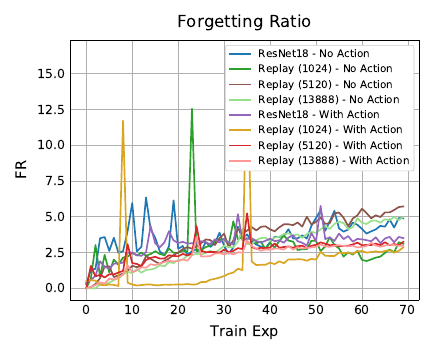}}
    \subfigure
    % {\includegraphics[width=0.33\textwidth]{img/replay_fwt.eps}}
    {\includegraphics[width=0.33\textwidth]{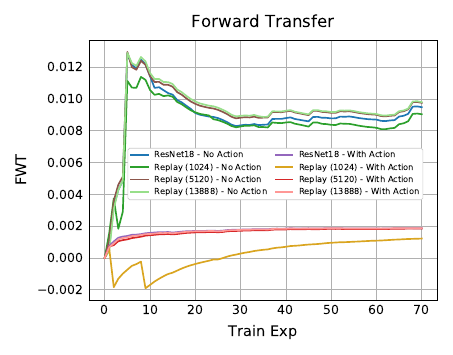}}
\end{center}
\vspace{-1em}
\caption{Continual metrics computed on replay with different buffer sizes. Small amounts of replay tend to slightly mitigate forgetting and even display positive backward transfer during an initial phase, but later converge to similar performance as naive or regularization techniques. The absence of forward transfer when action information is provided can be explained by the faster training convergence during the initial experiences (\cref{fig:avg_loss_action}).}
\label{fig:replay_bwt_fwt_fr}
\end{figure}
\begin{figure}[t]
\centering
\begin{minipage}{.47\textwidth}
  \centering  
  \includegraphics[width=0.9\textwidth]{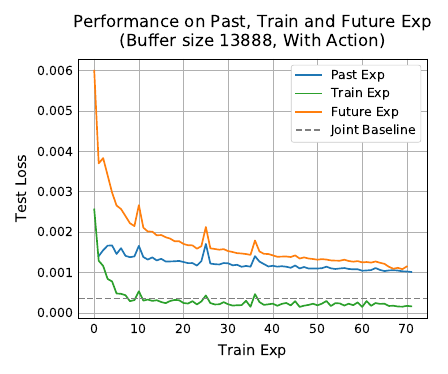}
  \captionof{figure}{Impact of replay (largest buffer size) when the model is given also action information.}
\label{fig:replay_act}
\end{minipage}%
\hfill
\begin{minipage}{.47\textwidth}
  \centering
\includegraphics[width=.95\textwidth]{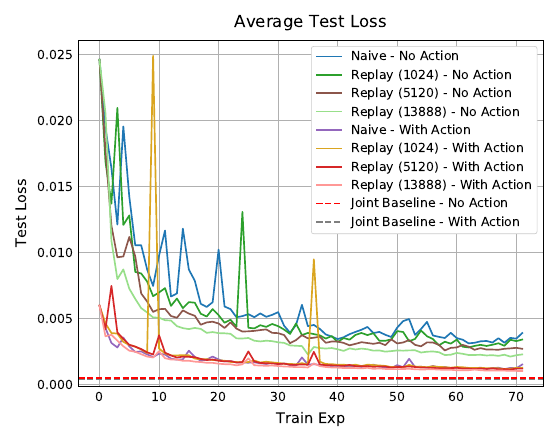}
\captionof{figure}{Comparison of the average test loss when adopting replay.}
\label{fig:action_replay_buffer_time}
\end{minipage}
\end{figure}
\begin{figure}[!htp]
\begin{center}
    \subfigure
    % {\includegraphics[width=0.47\textwidth]{img/2d_buffer_time.eps}}
    {\includegraphics[width=0.47\textwidth]{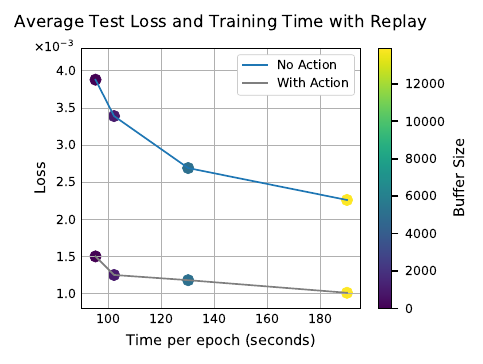}}
\end{center}
\vspace{-1.5em}
\caption{Relation between average test loss and training time for increasing replay buffer sizes, considering models trained without and with action information. Without action information, performance and training speed must be traded off. Contrarily, when the model is provided with action information, the curve flattens, suggesting that additional replay buffer capacity does not impact performance significantly.}
\label{fig:replay_buffer_time}
\end{figure}

\vspace{-1em}
\subsection{Impact of Action Information}

\begin{figure}[!htp]
\begin{center}
    \subfigure
    % {\includegraphics[width=0.32\textwidth]{img/actnaive_dz.eps}}
    {\includegraphics[width=0.32\textwidth]{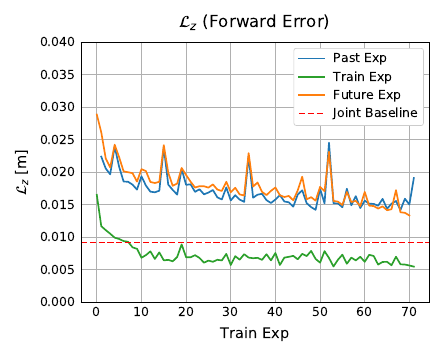}}
    \subfigure
    % {\includegraphics[width=0.32\textwidth]{img/actnaive_dx.eps}}
    {\includegraphics[width=0.32\textwidth]{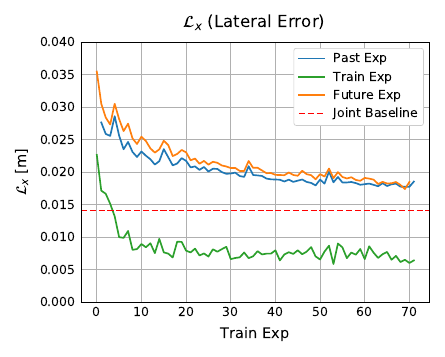}}
    \subfigure
    % {\includegraphics[width=0.31\textwidth]{img/actnaive_dyaw.eps}}
    {\includegraphics[width=0.31\textwidth]{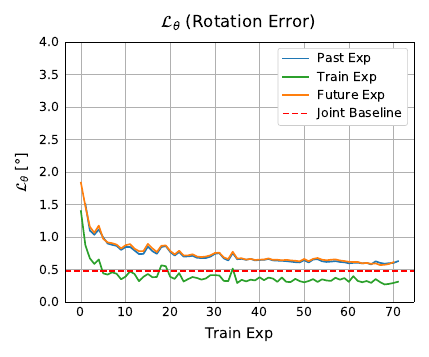}}
\end{center}
\vspace{-1.5em}
\caption{Comparison of each loss component over current, past, and future experiences, when action information is provided to the model.}
\label{fig:past_current_future_act}
\end{figure}

We present the loss profile over its three components when the model is provided action information (\Cref{fig:past_current_future_act}).

\FloatBarrier

\subsection{Impact of Model Scale}

We report additional evaluations referring to \cref{sec:impact_of_scale}, showing similar trends despite the added capacity.

\begin{figure}[!htp]
\begin{center}
    \subfigure
    % {\includegraphics[width=0.33\textwidth]{img/res50_all_bwt.eps}}
    {\includegraphics[width=0.33\textwidth]{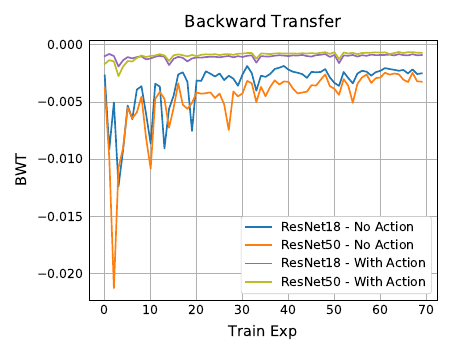}}
    \subfigure
    % {\includegraphics[width=0.31\textwidth]{img/res50_all_ratio.eps}}
    {\includegraphics[width=0.31\textwidth]{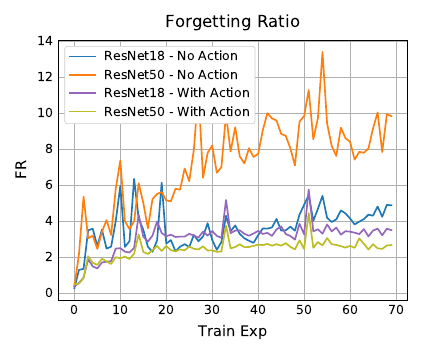}}
    \subfigure
    % {\includegraphics[width=0.32\textwidth]{img/res50_all_fwt.eps}}
    {\includegraphics[width=0.32\textwidth]{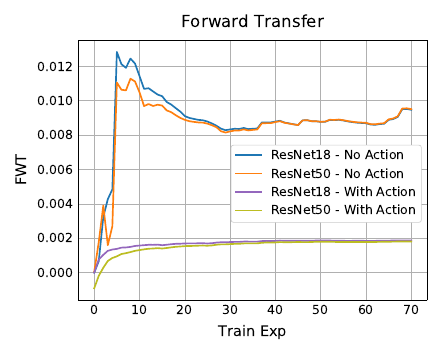}}
\end{center}
\vspace{-2em}
\caption{Impact of action information on continual metrics for the scaled Resnet50 model.}
\label{fig:res50_cl_metrics}
\end{figure}

\FloatBarrier
\vspace{-.5em}
\section{Detailed Experimental Results}
In \cref{tab:performance,tab:performance_resnet50,tab:performance_cl} we report a comprehensive summary of the experiments performed in this work.
\begin{table}[htp]
\centering
\begin{threeparttable}
    \caption{Intermediate and final performance of the tested algorithms in continual VO.}
    \label{tab:performance}
    \centering
    \begin{tabular}{lrrrrrr | SS }
        \toprule
        \textbf{Experience} & \textbf{0--11} & \textbf{12--23} & \textbf{24--35} & \textbf{36--47} & \textbf{48--59} & \textbf{60--71} & \textbf{Average}  & \textbf{Final}\\
        \midrule
        \multicolumn{9}{l}{\textit{No Action}}\\[.3em]
        Naive           & 13.75 & 7.25 & 5.01 & 3.93 & 4.06 & 3.36 & 6.23 & 3.88 \\
        EWC             & 13.13 & 6.24 & 4.80 & 3.87 & 3.70 & 3.39 & 6.05 & 4.29\\
        LwF             & 13.07 & 6.85 & 4.97 & 4.09 & 3.95 & 3.40 & 5.86 & 4.03 \\
        Replay (1024)   & 12.22 & 5.50 & 4.97 & 3.61 & 3.49 & 3.08 & 5.48 & 3.39 \\
        Replay (5120)   & 10.39 & 4.84 & 3.79 & 3.21 & 3.04 & 2.73 & 4.67 & 2.69 \\
        Replay (13888)  & 9.39 & 4.09 & 3.12 & 2.67 & 2.50 & 2.23 & 3.99 & 2.26 \\
        \multicolumn{8}{l}{Joint Baseline} & 0.50 \\
        \midrule
        \multicolumn{9}{l}{\textit{With Action}}\\[.3em]
        Naive           & 3.04 & 1.93 & 1.60 & 1.40 & 1.36 & 1.25 & 1.76 & 1.50 \\
        EWC             & 3.10 & 1.89 & 1.58 & 1.43 & 1.39 & 1.27 & 1.78 & 1.51 \\
        LwF             & 3.05 & 1.91 & 1.61 & 1.42 & 1.37 & 1.27 & 1.77 & 1.63 \\
        Replay (1024)   & 5.19 & 1.95 & 1.63 & 2.16 & 1.33 & 1.23 & 2.25  & 1.25\\
        Replay (5120)   & 3.69 & 1.92 & 1.65 & 1.49 & 1.28 & 1.18 & 1.87 & 1.18 \\
        Replay (13888) & 2.96 & 1.67 & 1.42 & 1.28 & 1.14 & 1.05 & 1.59 & 1.01 \\
        \multicolumn{8}{l}{Joint Baseline} & 0.35 \\
        \bottomrule
    \end{tabular}
    \begin{tablenotes}
      \small
      \item All loss values expressed in scale $\num{e-3}$.
    \end{tablenotes}
\end{threeparttable}
\end{table}
\begin{table}[htp]
\centering
\begin{threeparttable}
    \caption{Intermediate and final performance using the larger model ResNet50.}
    \label{tab:performance_resnet50}
    \centering
    \begin{tabular}{lrrrrrr | SS }
        \toprule
        \textbf{Experience} & \textbf{0--11} & \textbf{12--23} & \textbf{24--35} & \textbf{36--47} & \textbf{48--59} & \textbf{60--71} & \textbf{Average}  & \textbf{Final}\\
        \midrule
        \multicolumn{9}{l}{\textit{ResNet50 - No Action}}\\[.3em]
        Naive           & 16.19 & 8.69 & 6.83 & 5.17 & 4.69 & 3.87 & 7.57 & 4.62 \\
        \multicolumn{8}{l}{Joint Baseline} & 0.5 \\
        \midrule
        \multicolumn{9}{l}{\textit{ResNet50 - With Action}}\\[.3em]
        Naive           & 3.72 & 1.85 & 1.48 & 1.28 & 1.20 & 1.11 & 1.77 & 1.31 \\
        \multicolumn{8}{l}{Joint Baseline} & 0.23 \\
        \bottomrule
    \end{tabular}
    \begin{tablenotes}
      \small
      \item All loss values expressed in scale $\num{e-3}$.
    \end{tablenotes}
\end{threeparttable}
\end{table}
\begin{table}[!htp]
\centering
\begin{threeparttable}
    \caption{Continual learning metrics.}
    \label{tab:performance_cl}
    \centering
    \begin{tabular}{l SS SS SS}
        \toprule
        \textbf{} & \multicolumn{2}{c}{\textbf{BWT}$^{*}$ } & \multicolumn{2}{c}{\textbf{FR}} & \multicolumn{2}{c}{\textbf{FWT}$^{*}$} \\
                  & \textbf{Average} & \textbf{Final} & \textbf{Average} & \textbf{Final} & \textbf{Average} & \textbf{Final} \\
        \midrule
        \multicolumn{4}{l}{\textit{No Action}}\\[.3em]
        Naive           & -3.48 & -2.50 & 3.60 & 4.88 & 8.79 & 9.48 \\
        EWC             & -3.27 & -2.53 & 3.66 & 5.03 & 9.14 & 9.66 \\
        LwF             & -3.72 & -2.81 & 4.05 & 5.25 & 8.96 & 9.61 \\
        Replay (1024)   & -2.17 & -1.85 & 2.80 & 3.11 & 8.40 & 9.03 \\
        Replay (5120)   & -1.67 & -2.07 & 3.46 & 5.72 & 9.14 & 9.75 \\
        Replay (13888)  & -0.74 & -1.54 & 2.89 & 4.98 & 9.21 & 9.80 \\
        \midrule
        \multicolumn{4}{l}{\textit{With Action}}\\[.3em]
        Naive           & -1.04 & -0.90 & 3.10 & 3.51 & 1.70 & 1.87 \\
        EWC             & -1.04 & -0.94 & 3.12 & 3.77 & 1.68 & 1.87 \\
        LwF             & -1.04 & -0.91 & 3.11 & 3.62 & 1.68 & 1.86 \\
        Replay (1024)   & -0.15 & -0.26 & 1.70 & 2.86 & 0.25 & 1.21 \\
        Replay (5120)   & -0.93 & -0.84 & 2.54 & 3.25 & 1.62 & 1.84 \\
        Replay (13888)  & -0.63 & -0.70 & 2.27 & 2.97 & 1.68 & 1.87 \\
        \bottomrule
    \end{tabular}
    \begin{tablenotes}
      \small
      \item $^{*}$Values expressed in scale $\num{e-3}$.
    \end{tablenotes}
\end{threeparttable}
\end{table}

\FloatBarrier
\clearpage
\section{Details on dataset composition}
\label{sec:dataset}

Our dataset focuses on scene-level continual granularity. For each of the 72 train apartments, we sample several trajectories and accumulate the observations we obtained. In doing so, we also make sure that each action is represented in a balanced manner, and we guarantee that collisions with environment features (walls and obstacles) are captured enough times, in order for the model to observe them enough. \Cref{tab:cldataset} shows these details on the collected samples, while  \cref{tab:clproportion} contains details on the number of samples used in our train, validation, and test splits.

\begin{figure}[htp]
\begin{center}
    \subfigure
    {\includegraphics[width=0.27\textwidth]{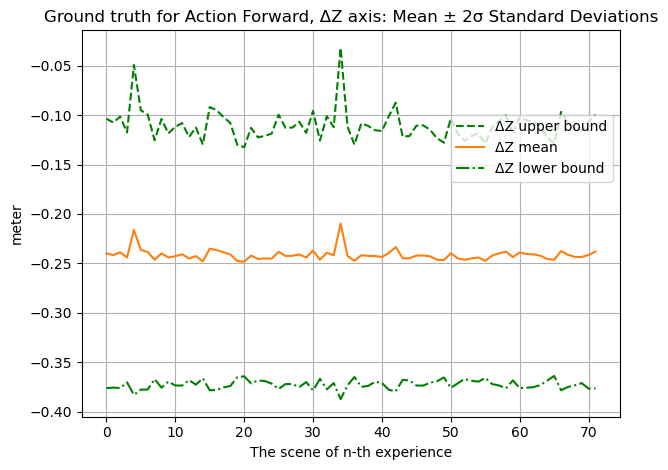}}
    \subfigure
    {\includegraphics[width=0.27\textwidth]{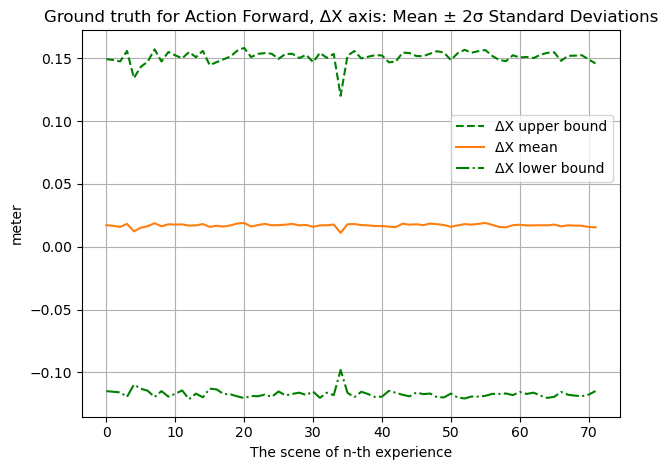}}
    \subfigure
    {\includegraphics[width=0.27\textwidth]{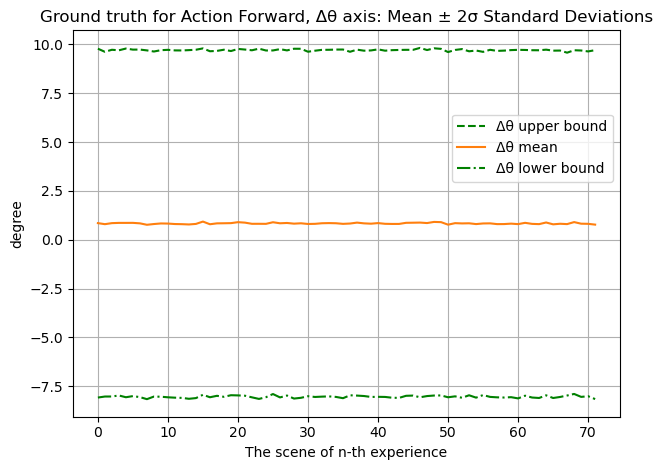}} \\
    \subfigure
    {\includegraphics[width=0.27\textwidth]{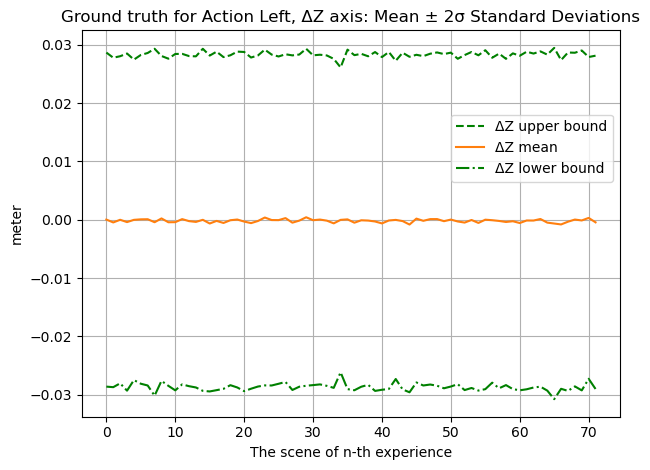}}
    \subfigure
    {\includegraphics[width=0.27\textwidth]{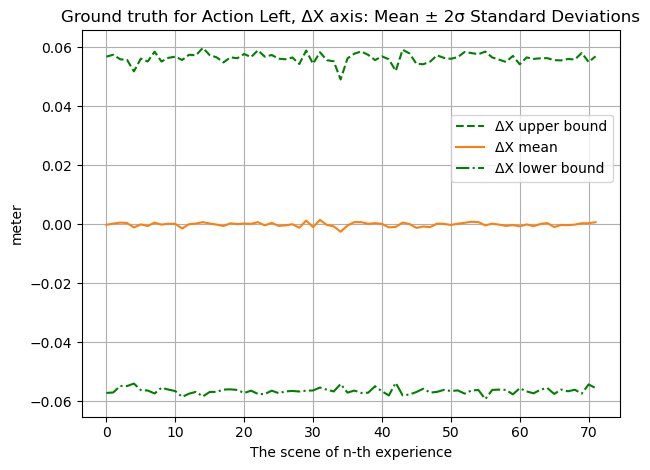}}
    \subfigure
    {\includegraphics[width=0.27\textwidth]{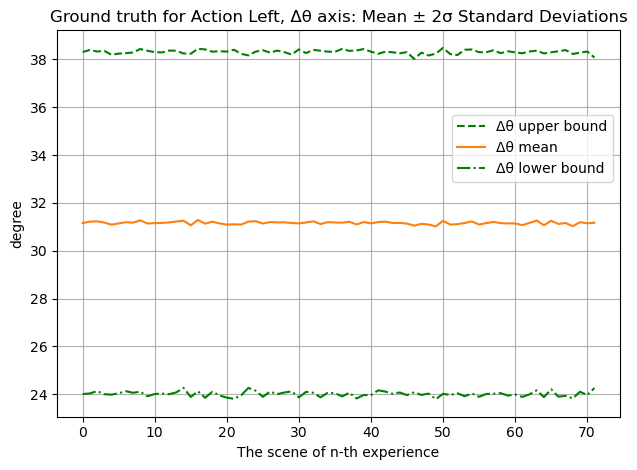}} \\
    \subfigure
    {\includegraphics[width=0.27\textwidth]{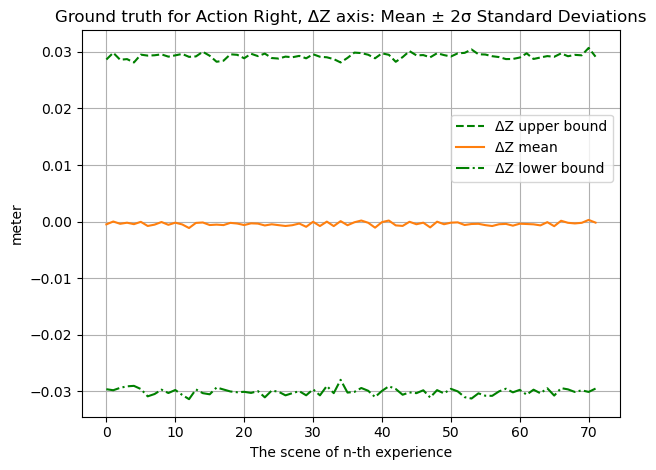}}
    \subfigure
    {\includegraphics[width=0.27\textwidth]{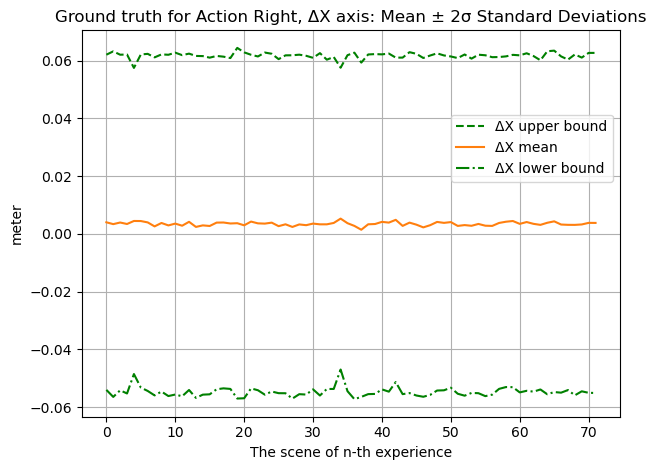}}
    \subfigure
    {\includegraphics[width=0.27\textwidth]{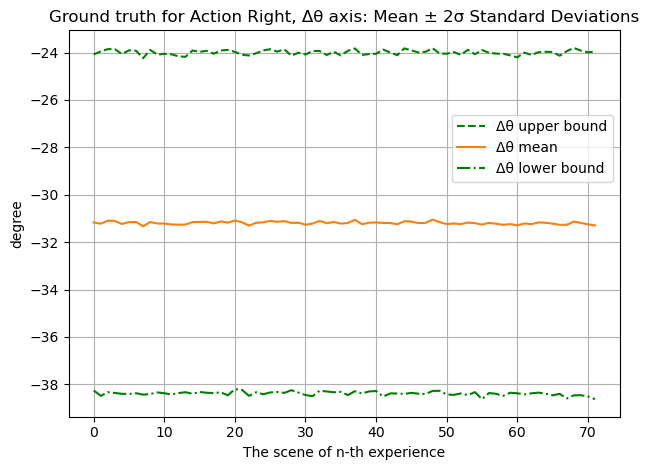}}
\end{center}
\vspace{-1em}
\caption{Ground truth output variability for each action, when computed from all experiences in our dataset.}
\label{fig:gt_ci}
\end{figure}

In \cref{fig:gt_ci} we analyze, for reference, the distribution of the motion resulting from each action, approximated with an interval of two standard deviations around its mean.

\begin{table}[htp]
    \caption{Statistics of collected dataset}
    \label{tab:cldataset}
    \centering
    \begin{tabular}{lrrrr}
        \toprule
        \textbf{Action} & \textbf{\# Frames} & \textbf{Frames (\%)} & \textbf{\# Collisions} & \textbf{Collisions (\%)} \\
        \midrule
        Forward & 700803 & 57.39 & 83261 & 11.88\\
        Left & 260559 & 21.34 & 32999 & 12.66\\
        Right & 259758 & 21.27 & 19769 & 7.61\\
        \midrule
        Overall     & 1221120 & 100.00 & 136029 & 11.14  \\
        \bottomrule
    \end{tabular}
\end{table}
\begin{table}[h]
    \centering
    \caption{Details of the dataset split.}
    \label{tab:clproportion}
    \begin{tabular}{ccc}
        \toprule
        \textbf{Dataset} & \textbf{Total Count} & \textbf{Total Proportion} \\
        \midrule
        Training & 999936 & 81.89 \% \\
        Test & 147456 & 12.08 \% \\
        Validation & 73728 & 6.04 \% \\
        \bottomrule
    \end{tabular}
\end{table}

\FloatBarrier
\clearpage
\section{Model Prediction Variability}
\label{sec:action_prediction_variability}

\begin{figure}[htp]
\begin{center}
    \subfigure
    % {\includegraphics[width=0.3\textwidth]{img/MCI_ResNet18_Forward_Z.eps}}
    {\includegraphics[width=0.3\textwidth]{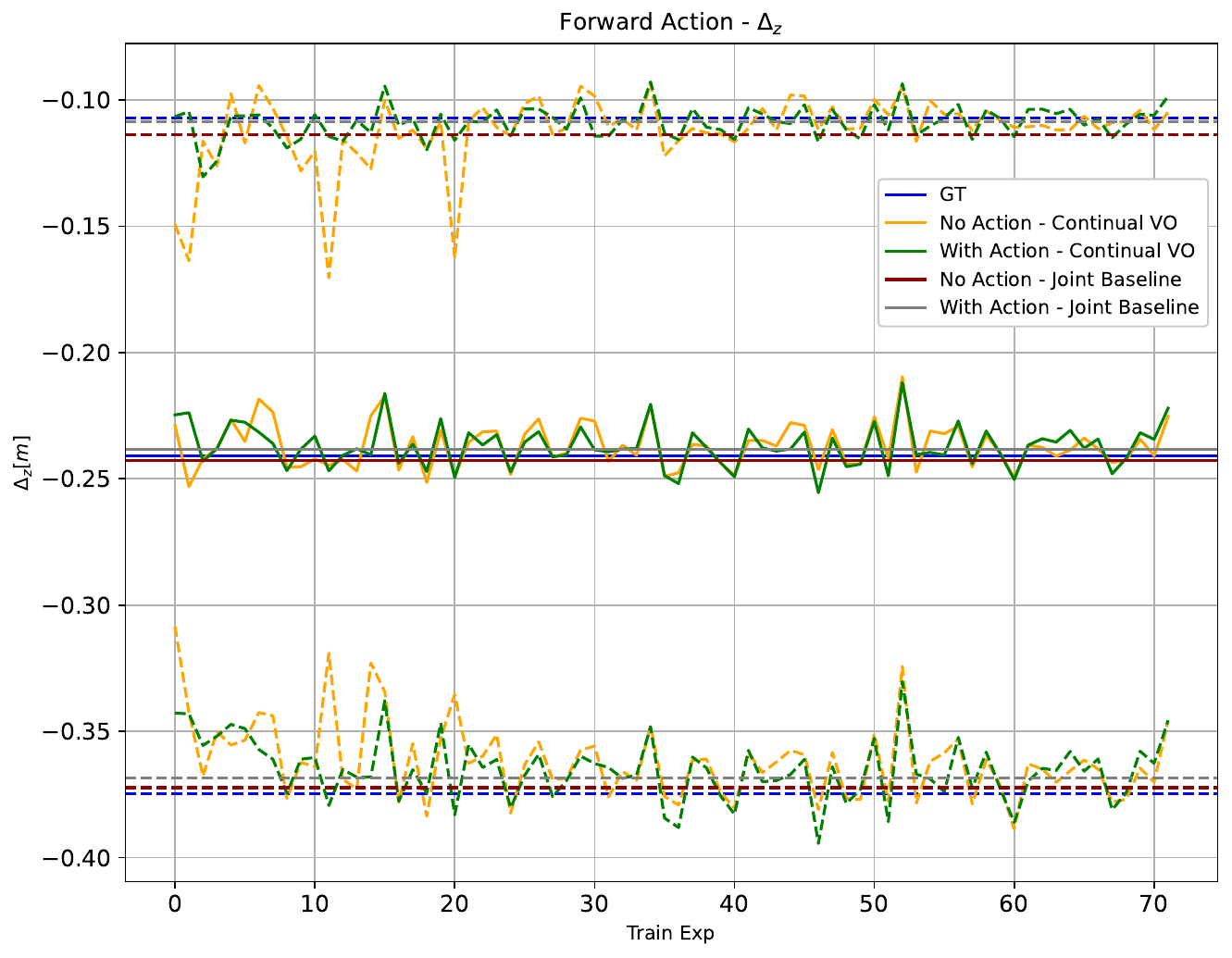}}
    \subfigure
    % {\includegraphics[width=0.3\textwidth]{img/MCI_ResNet18_Forward_X.eps}}
    {\includegraphics[width=0.3\textwidth]{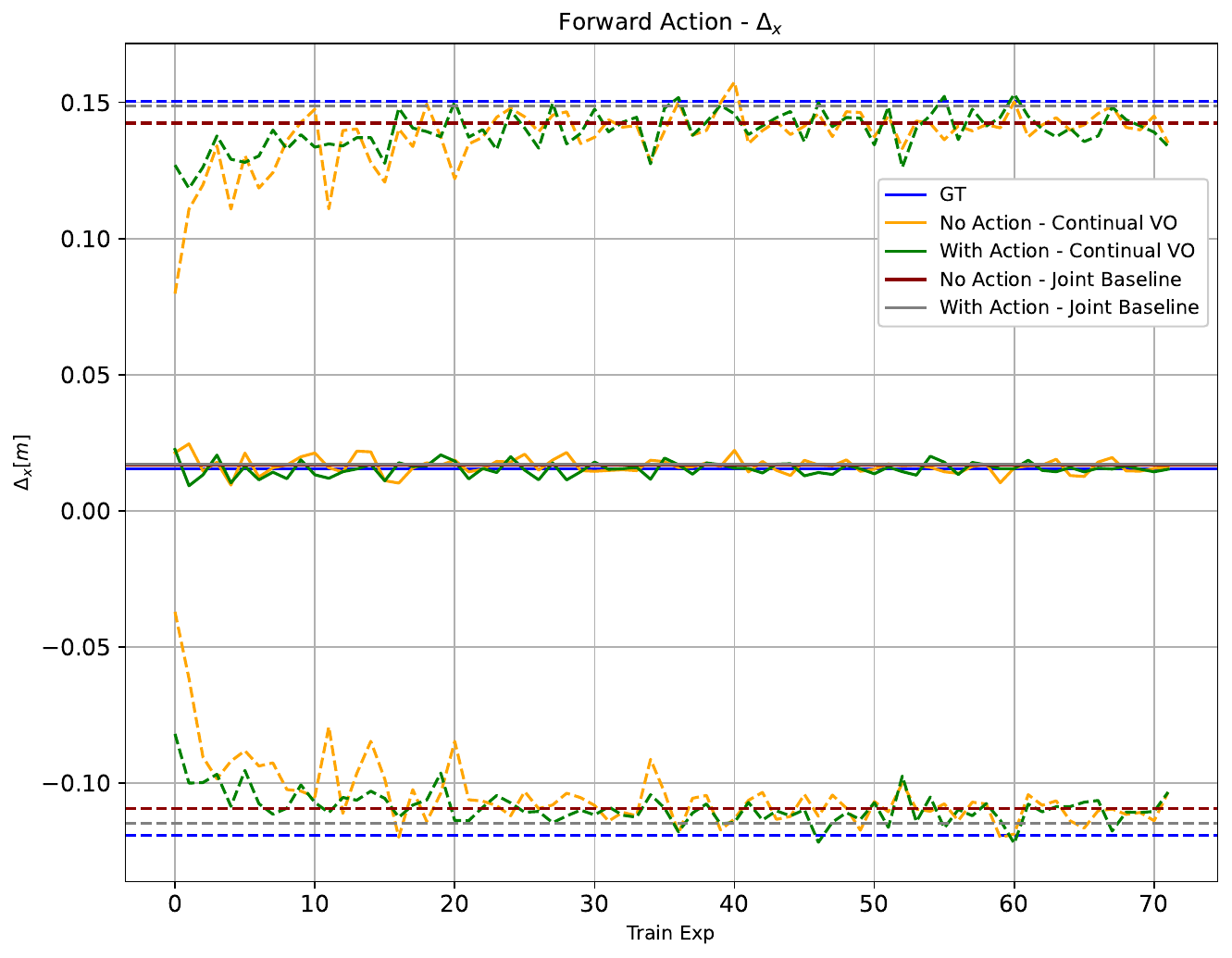}}
    \subfigure
    % {\includegraphics[width=0.3\textwidth]{img/MCI_ResNet18_Forward_YAW.eps}} \\
    {\includegraphics[width=0.3\textwidth]{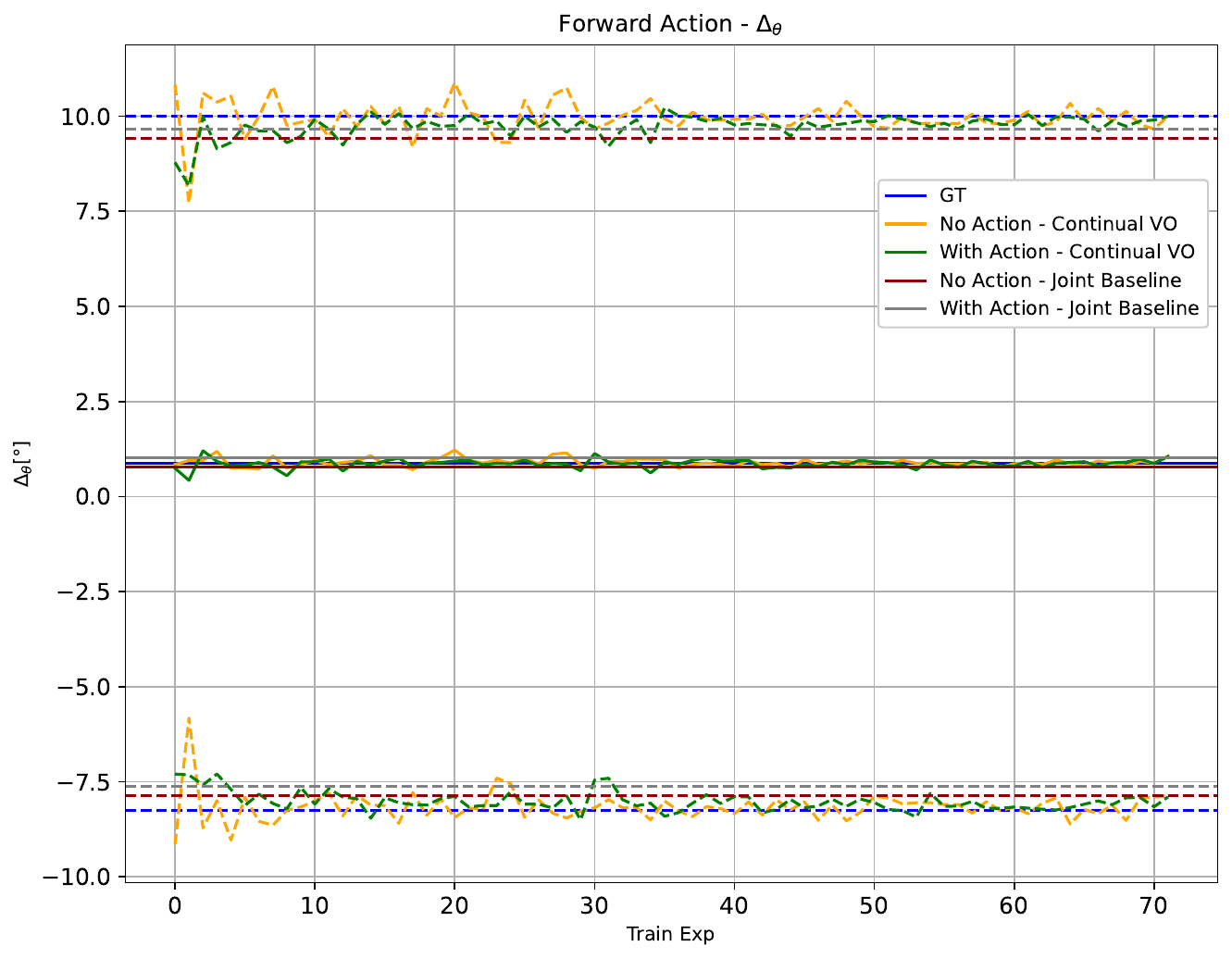}} \\
    \subfigure
    % {\includegraphics[width=0.3\textwidth]{img/MCI_ResNet18_Left_Z.eps}}
    {\includegraphics[width=0.3\textwidth]{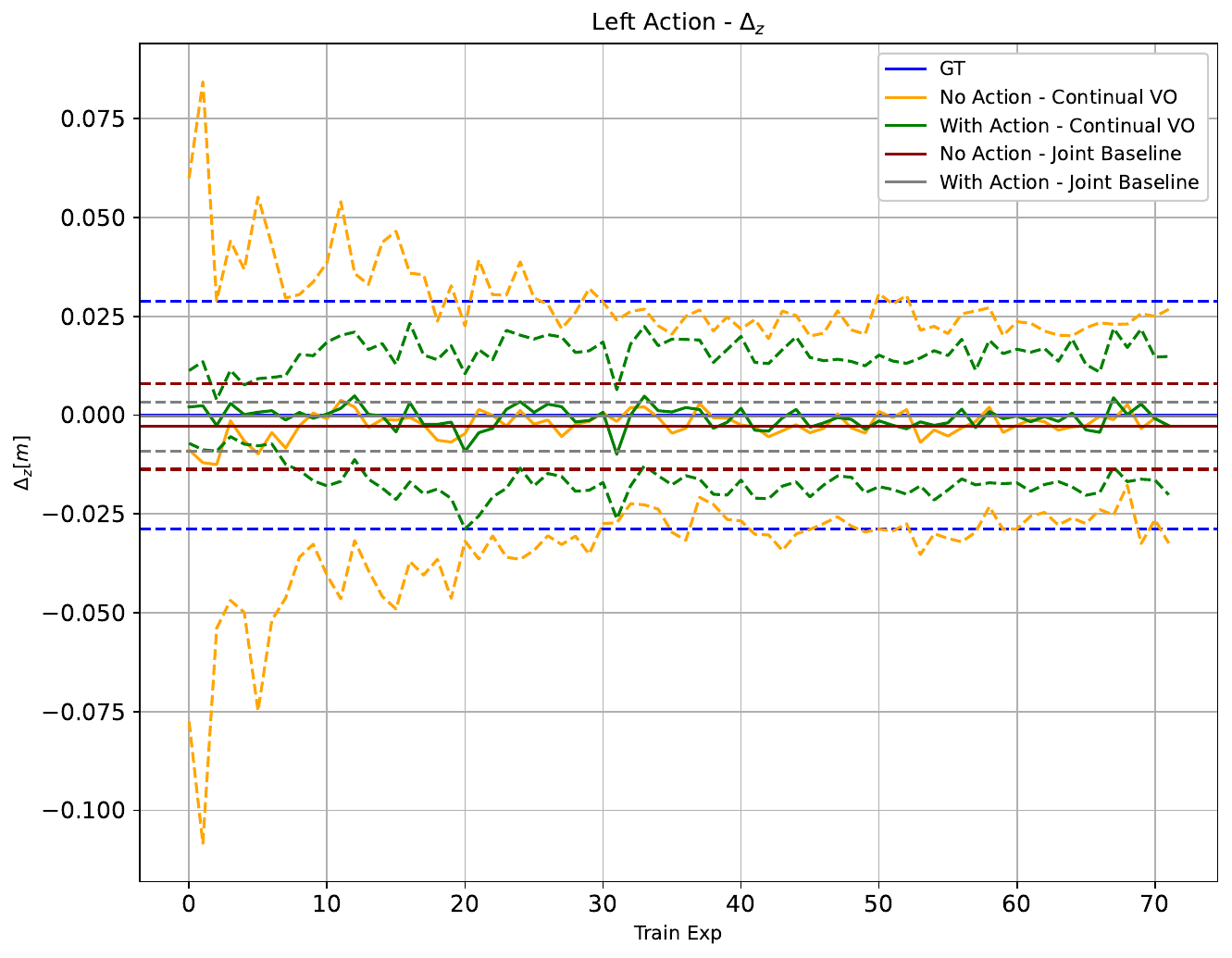}}
    \subfigure
    % {\includegraphics[width=0.3\textwidth]{img/MCI_ResNet18_Left_X.eps}}
    {\includegraphics[width=0.3\textwidth]{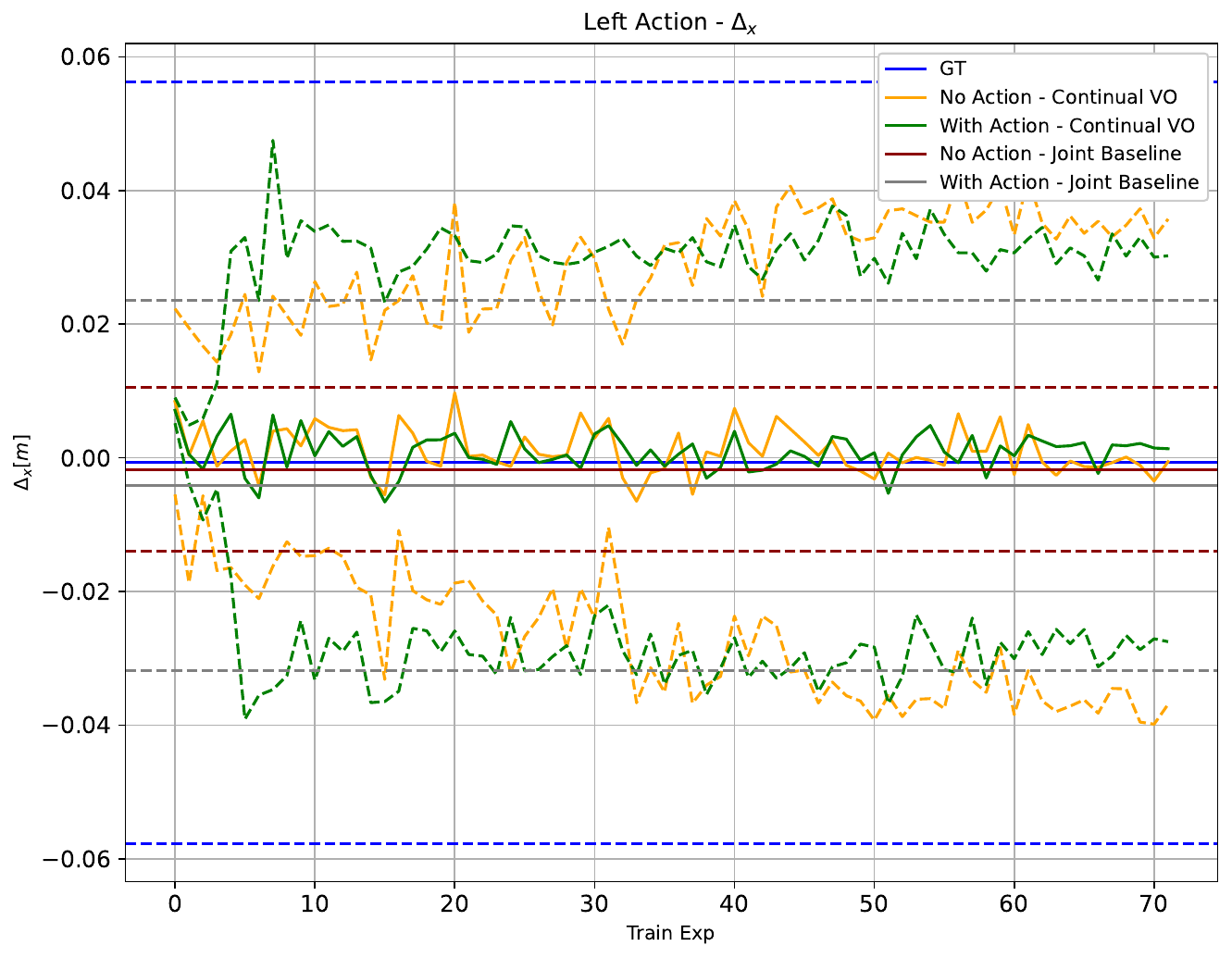}}
    \subfigure
    % {\includegraphics[width=0.3\textwidth]{img/MCI_ResNet18_Left_YAW.eps}} \\
    {\includegraphics[width=0.3\textwidth]{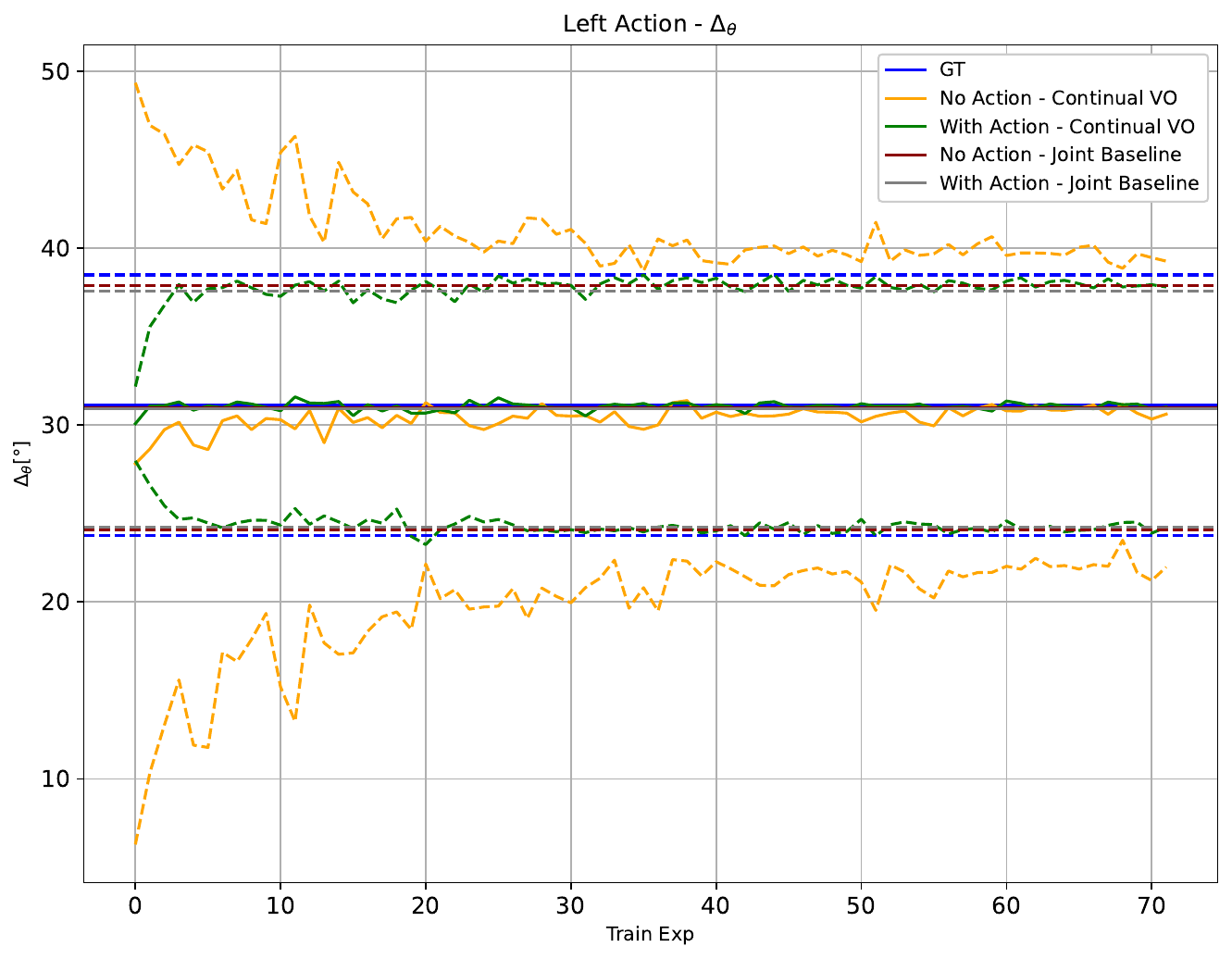}} \\
    \subfigure
    % {\includegraphics[width=0.3\textwidth]{img/MCI_ResNet18_Right_Z.eps}}
    {\includegraphics[width=0.3\textwidth]{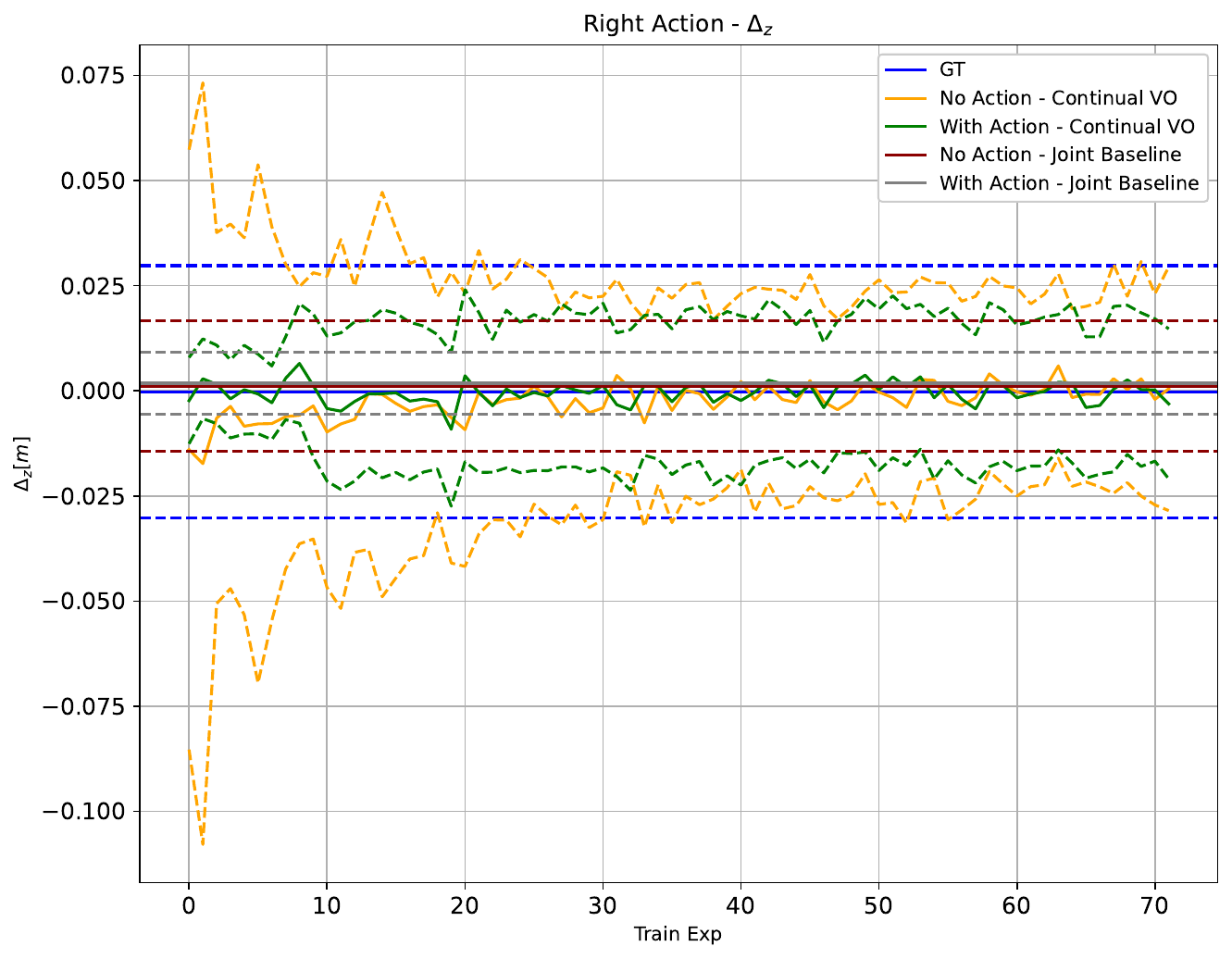}}
    \subfigure
    % {\includegraphics[width=0.3\textwidth]{img/MCI_ResNet18_Right_X.eps}}
    {\includegraphics[width=0.3\textwidth]{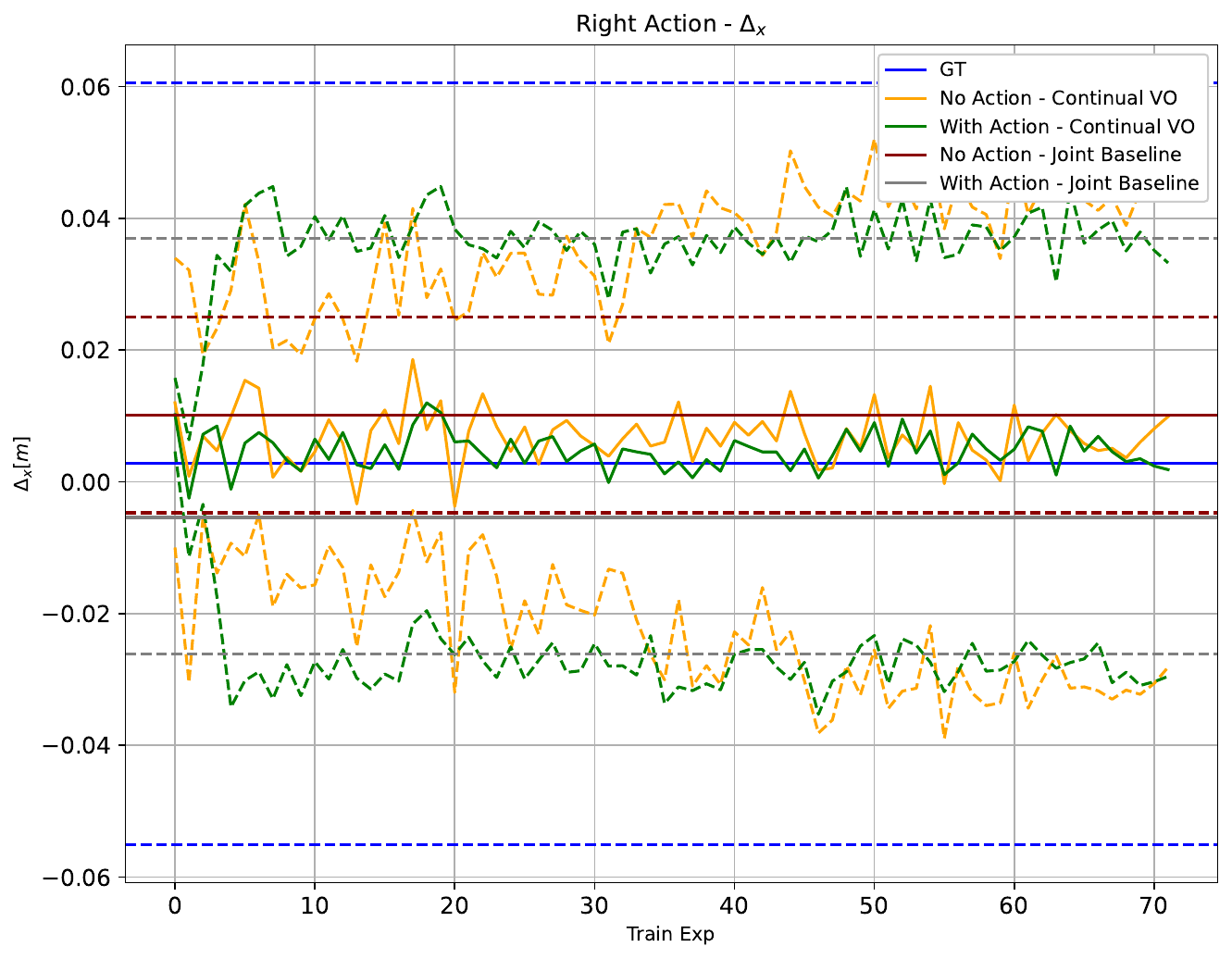}}
    \subfigure
    % {\includegraphics[width=0.3\textwidth]{img/MCI_ResNet18_Right_YAW.eps}}
    {\includegraphics[width=0.3\textwidth]{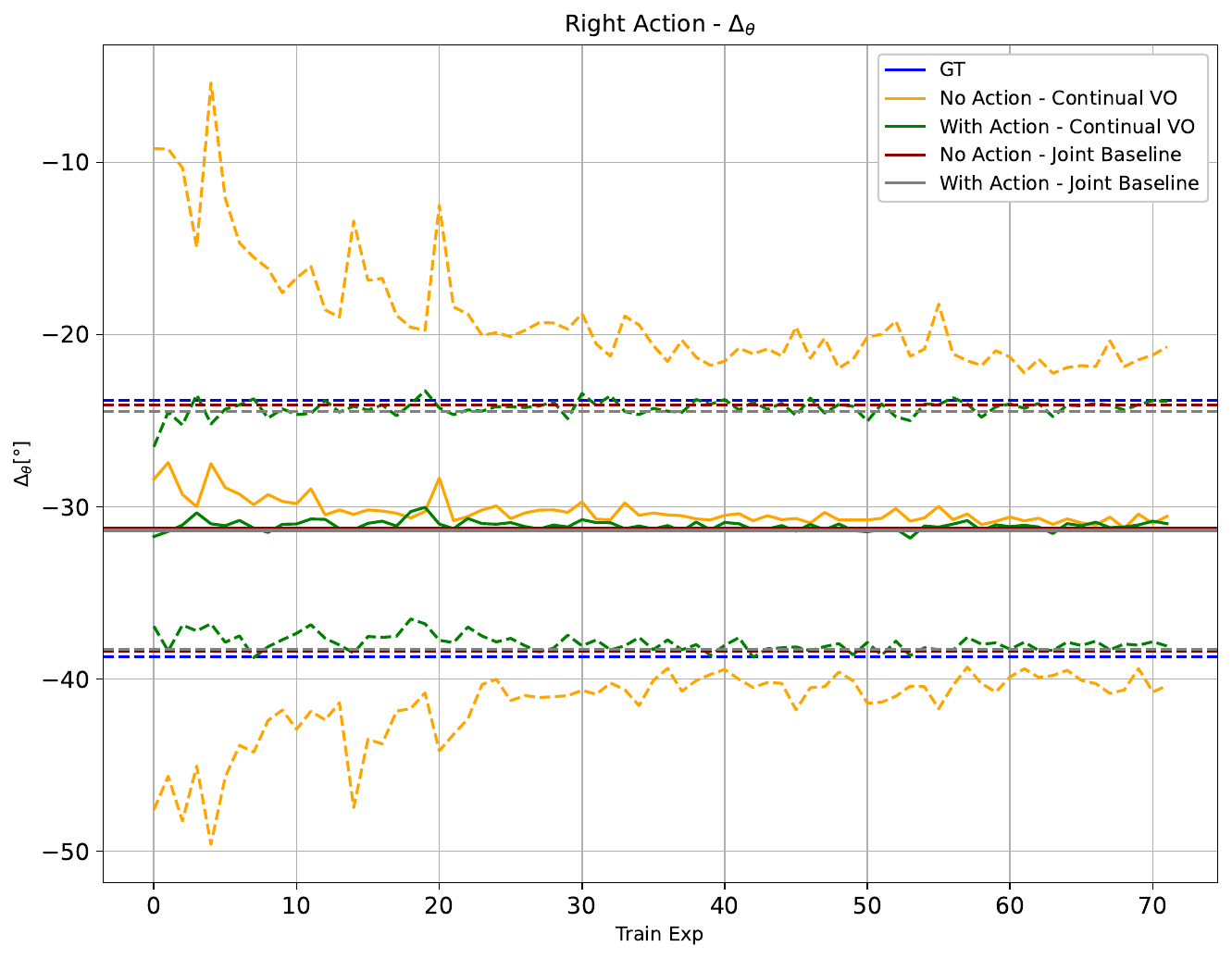}}
\end{center}
\caption{Prediction variability of the models.}
\label{fig:pred_var_act}
\end{figure}

We perform an additional analysis of the model behavior when trained without or with action information, studying the distribution of the output values produced throughout its life.
In particular, we approximate such distribution as Gaussian and display an interval of two standard deviations from the mean, computed after applying the model to the entire test set of each experience.
We plot this temporal output dynamics in \cref{fig:pred_var_act}.
For reference, the mean and standard deviation of the ground truth are displayed as well.

The trends show interesting properties.
When the action information is provided to the model, it very quickly learns to predict the mean value for each action distribution, with very low variance. With time, it starts integrating the information received from the images, and thus predicting values farther from the mean, slowly matching the ground truth distribution.

The opposite happens to the model relying only on visual information. fact, it initially tends to estimate displacements with high variance, producing an output distribution much more dispersed than required. This is because the model is learning at the same time to reproduce all three very different distributions simultaneously. This dispersion is reduced throughout its lifetime, but never completely lost, and after all experiences its output distribution is still wider than the ground truth.

\end{document}